\documentclass[10pt,twocolumn,letterpaper]{article}

\usepackage[pagenumbers]{cvpr} 

\definecolor{cvprblue}{rgb}{0.21,0.49,0.74}
\definecolor{linkcolor}{RGB}{255,0,0}
\definecolor{urlcolor}{RGB}{255,105,180}
\usepackage[pagebackref,breaklinks,colorlinks=true,allcolors=cvprblue,linkcolor=linkcolor,urlcolor=urlcolor]{hyperref}
\usepackage{multirow} 
\usepackage{microtype}      
\usepackage{xcolor}         
\usepackage{graphicx}
\usepackage{amsmath}
\usepackage{subcaption}
\usepackage{arydshln}
\usepackage{wrapfig}
\usepackage{marvosym}

\usepackage{booktabs}
\newcommand{\best}[1]{\textcolor{red}{#1}}
\newcommand{\second}[1]{\textcolor{blue}{#1}}
 
\title{Exploring Real\&Synthetic Dataset and Linear Attention in Image Restoration}

\author{
Yuzhen Du$^{1\ast}$
  ~~ Teng Hu$^{1}$\thanks{Equal contribution.~~~~~~Jiangning Zhang leads this project.}
  ~~ Jiangning Zhang$^{2,3}$
  ~~ Ran Yi$^{1}$\thanks{Corresponding author.}
  ~~ Chengming Xu$^2$ \\
  ~~ Xiaobin Hu$^2$
  ~~ {Kai Wu}$^2$
  ~~ {Donghao Luo}$^2$
  ~~ {Yabiao Wang}$^2$
  ~~ {Lizhuang Ma}$^1$\\
  \normalsize $^1$Shanghai Jiao Tong University ~~ $^2$Youtu Lab, Tencent ~~ $^3$Zhejiang University \\
  \tt\small Code: \url{https://github.com/YuzhenD/Resyn}
}

\begin{document}
\maketitle
\begin{abstract}
Image restoration (IR) aims to recover high-quality images from degraded inputs. Recent advancements in deep learning have significantly improved image restoration performance. However, existing methods lack a unified training benchmark specifying training iterations and configurations. Additionally, we construct an image complexity evaluation metric using the gray-level co-occurrence matrix (GLCM) and find a bias between the image complexity distributions of commonly used IR training and testing datasets, leading to suboptimal restoration results. Therefore, we construct a new large-scale IR dataset called ReSyn, that utilizes a novel image filtering method based on image complexity to achieve a balanced image complexity distribution and contains both real and AIGC synthetic images. From the perspective of measuring the model's convergence ability and restoration capability, we construct a unified training standard that specifies the training iterations and configurations for image restoration models.  Furthermore, we explore how to enhance the performance of transformer-based image restoration models based on linear attention mechanisms. We propose \textbf{RWKV-IR}, a novel image restoration model that incorporates the linear complexity RWKV into the transformer-based image restoration structure, and enables both global and local receptive fields. Instead of directly integrating Vision-RWKV into the transformer architecture, we replace the original Q-Shift in RWKV with a novel Depth-wise Convolution shift to effectively model the local dependencies. It is further combined with Bi-directional attention to achieve both global and local aware linear attention. Moreover, we propose a Cross-Bi-WKV module that combines two Bi-WKV modules with different scanning orders to achieve balanced attention for horizontal and vertical directions. Extensive quantitative and qualitative experiments demonstrate the effectiveness of our RWKV-IR model. Project: \url{https://yuzhend.github.io/ReSyn.github.io/}

\end{abstract}    
\section{Introduction}
\label{sec:intro}
Image restoration (IR), which aims to recover high-quality images from low-quality degraded inputs, is {a} crucial {task} in modern image processing. This field encompasses various sub-tasks, including super-resolution, image denoising, and compression artifact{s} reduction. Recently, the {advancements} of deep learning techniques, such as Convolutional Neural Networks (CNNs)~\cite{dai2019second,dong2014learning,lim2017enhanced,DnCNN,zhang2018image} and Transformers~\cite{chen2021pre,chen2023activating,chen2023dual,li2023grl,liang2021swinir}, {have} significantly enhanced image restoration performance, 
driving continuous progress in {this} field. 
Reviewing the previous IR methods, more complex and deeper models~\cite{zhou2023srformer, chen2023activating} often achieve better performance. 

\begin{figure*}[t]
     \centering
     \includegraphics[width=0.98\textwidth]{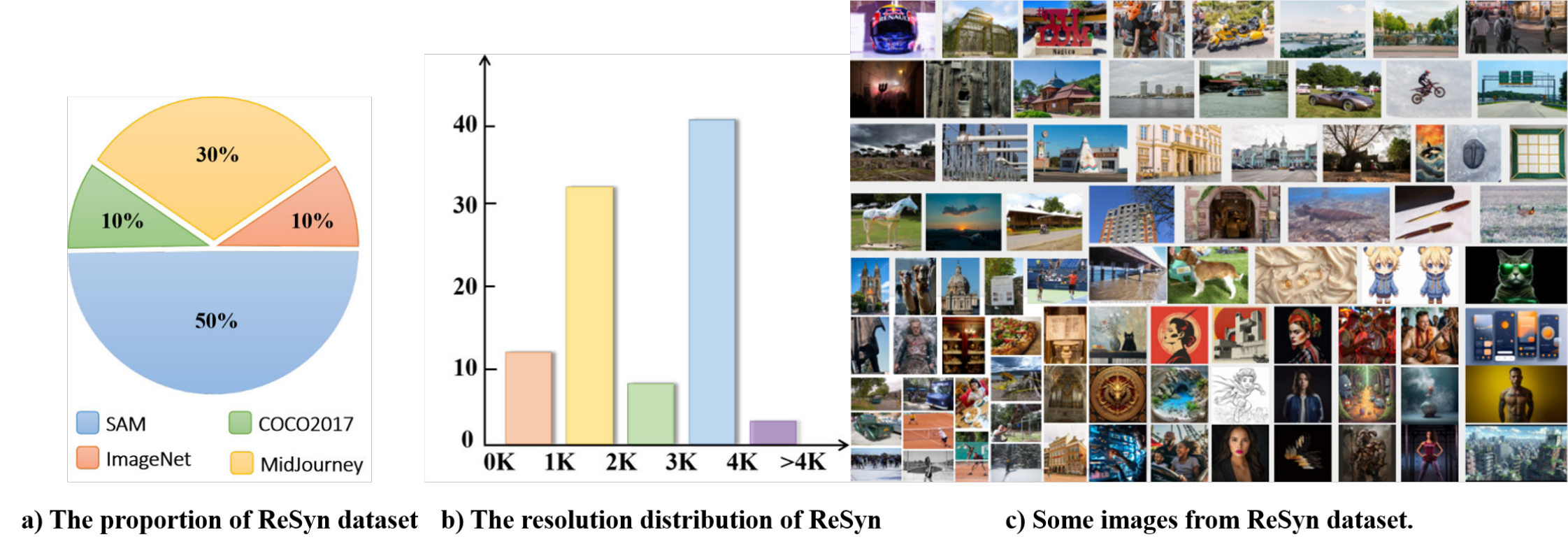}
     \caption{The diversity analysis of our ReSyn dataset. It {contains both real and synthetic images from a variety of} data sources and covers a wide range of resolutions. }
     \label{fig: Diversity}
\end{figure*}

{To meet the data requirements for {IR} model training, a large number of images need to be collected to {construct} a paired {training} dataset. Due to {limited photography and compression techniques}, the images {in the previous training datasets} often have the problem of blur{ring} or noise, {meanwhile, the images in some test datasets have more image details,} this causes the domain gap {(different image complexities)} between the common{ly} used training and test {datasets}~\cite{li2023lsdir}. 
Most datasets~\cite{timofte2017ntire,li2023lsdir} focused on collecting a large number of images of high resolution, few consider how to measure {and address} this {domain gap}.}

In this paper, we construct an image complexity metric based on the Gray-Level Co-occurrence Matrix (GLCM) analysis method{, and} reveal a significant bias in image complexity distribution within classic IR training datasets and test datasets. We further utilize this metric as a criterion for filtering the image restoration dataset and construct {a new} image restoration dataset{,} which is called \textbf{ReSyn dataset}. 
In many previous datasets~\cite{li2023lsdir, timofte2017ntire}, resolution {and Bits Per Pixel (BPP) have} been {used as} important {criteria} for image filtering{, but they are insufficient for constructing \textit{image complexity balanced datasets}. To this end,}
we further utilize the {GLCM-based image complexity} metric we proposed to filter and retain some images of medium resolution but with high image complexity. 
{Moreover}, with the rapid development of AI-generated content (AIGC), there is a surging demand for synthetic image restoration. We consider the generated images as an essential part of the dataset and filter them in the same manner{, which} also enriches the sources of our dataset. 
The final ReSyn dataset comprises 12,000 images, with 30\% being high-quality synthetic images sourced from the web. {Our ReSyn dataset} also presents a wide {range} of image resolutions, ranging from 0.25K to 4K, and originates from various sources. 

We also review the training processes of previous image restoration models and find that there is a lack of a \textbf{unified {IR} training benchmark}{, \textit{i.e.,} the training iterations and configurations are not unified}. Considering the model's convergence and restoration capability, we construct a set of training {standards}. 
To measure the model's convergence capability, we use a shorter number of training iterations; to measure the restoration capability, we use a longer number of training iterations. This combination of short and long training iterations allows users to have a more comprehensive understanding of the model's capabilities, facilitating the selection of the model. 
{We conduct a comprehensive evaluation of SOTA IR models using the unified training standard, on our ReSyn dataset and other commonly used IR datasets.}
Both the ReSyn dataset and the {constructed unified} training {standard} form our proposed benchmark.

{When comparing various image restoration models, we notice that the linear attention mechanism {(\textit{e.g.,} Mamba-based IR models)} has great potential for enhancing the effective receptive field of models.} Therefore, we {aim} to {incoperate a} linear complexity attention mechanism, RWKV~\cite{peng2023rwkv}, with the image restoration models. 
{Based on the common{ly} used {transformer-based IR} structure, we construct our \textbf{RWKV-IR}, which consists of three stages: shallow feature extraction, deep-feature enhancing, and HQ image reconstruction modules.}
{We mainly incorporate RWKV into the second stage, deep-feature enhancing.}
We first introduce the Vision RWKV module to {extract} the deep image features, where a Spatial Mix Layer is employed to enable our model with global receptive fields with only linear {computational} complexity. 
Then, to enhance the {modeling of relationships in} local receptive fields and eliminate the negative effects caused by {the} original Q-shift {operation}, we exploit the characteristics of the local receptive field{s} in convolutional operations, {and propose} {a \textbf{{Depth-wise Convolution Shift (DC-shift)}} module,} 
as a replacement of the original Q-shift of the RWKV module. {Moreover}, the Bi-WKV method of origin{al} Vision-RWKV has an unbalanced position embedding, {paying more attention to the horizontal direction and less attention to the vertical direction,} which is not suitable for the IR tasks. 
We {propose} a \textbf{Cross-Bi-WKV module}, which {combines} two Bi-WKV {modules with different scanning orders} to achieve {a balanced} attention to surrounding features.
By cross-scanning and synchronizing the calculations of the two Bi-WKV modules in both horizontal and vertical directions, {we} achieve {a} balanced attention to all four directions.
Extensive quantitative and qualitative experiments demonstrate the effectiveness of our RWKV-IR. We summarize our contributions as below:
\begin{itemize}
    \item We propose a comprehensive benchmark for image restoration tasks, which includes a novel large-scale benchmark dataset, and a {unified} training standard that specifies the number of training iterations and the configuration of batch size. {We conduct a comprehensive evaluation of SOTA IR models using the unified training standard on the new dataset and other common IR datasets.}
    \item We construct a large-scale dataset called ReSyn, that integrates {both} real and synthetic images. This dataset encompasses a variety of {data} sources and utilizes novel image filter methods based on our {newly} proposed {GLCM-based} image complexity metric. 
    \item We design RWKV-IR, a novel image restoration model with both global and local receptive fields, which can effectively restore low-quality images with linear computational complexity. {We} replace {the original} token-shift {method (Q-shift)} with Depth-wise Convolution {shift for local dependencies modeling,} and {proposes Cross-Bi-WKV to replace} Bi-WKV {for more balanced attention for horizontal and vertical directions}, which enables RWKV to be effectively transferred to IR models.
\end{itemize}
\section{Related Works} \label{section:related}

\subsection{Image Restoration}
Image restoration has witnessed significant progress with the advent of computer vision~\cite{liang2021swinir,scsnet,cmos,guo2024mambair}, exemplified by pioneering CNN-based methods like SRCNN~\cite{dong2014learning}, DnCNN~\cite{DnCNN}, and ARCNN~\cite{dong2015compression}, targeting super-resolution, denoising, and artifact reduction~\cite{kim2016accurate,zhang2021plug,cavigelli2017cas,wang2018esrgan,zhang2018residual,lai2017deep,wei2021unsupervised,fu2019jpeg,zhang2018rcan,dai2019second}. Despite their success, CNN-based approaches often struggle to model global dependencies effectively. Meanwhile, Transformers, proven competitors to CNNs in various computer vision tasks~\cite{carion2020end,dosovitskiy2020image,liu2021swin,eat,emo}, show promise in restoration tasks. However, they encounter challenges due to the quadratic computational complexity of the attention mechanism~\cite{vaswani2017attention}. Strategies like IPT~\cite{chen2021pre} and SwinIR~\cite{liang2021swinir} address this by employing patch-based processing and shifted window attention. But the trade-off persists between efficient computation and global modeling~\cite{zhang2023accurate,chen2023activating,li2021efficient,chen2023dual,zamir2022restormer,chen2023recursive}. Recently, MambaIR~\cite{guo2024mambair} has been proposed to incorporate Mamba~\cite{gu2023mamba,liu2024vmamba} into the image restoration task, which globally processes the image features with only linear computational complexity. Following this trend, this paper explores the possibility of integrating another linear attention mechanism, RWKV (which has shown better performance in other vision tasks~\cite{fei2024diffusion,he2024pointrwkv,gu2024rwkv}), into image restoration models. 

\paragraph{Single Image Restoration Datasets.} 
Learning-based image restoration methods rely on the external training dataset to learn the mapping between degraded and GT images. But most training datasets~\cite{timofte2017ntire, li2023lsdir} focus on collecting higher resolutions and larger quantities of real images, few considering the bias between the training and testing datasets, and there is no dataset taking the synthetic images into account. In this paper, we construct an image complexity metric based on GLCM analysis and find that the commonly used training datasets~\cite{timofte2017ntire,lim2017enhanced} for SR task and~\cite{arbelaez2010contour,ma2016waterloo} for image denoising task, exhibit a certain distribution difference in resolution compared to the testing datasets~\cite{bevilacqua2012low,zeyde2012single,martin2001database,huang2015single,matsui2017sketch}. Therefore, we utilize this metric as a filter criterion and construct a new image ReStoration dataset which including the Real and Synthetic images.

\begin{figure}[t]
     \centering
     \centering
     \includegraphics[width=0.47\textwidth]{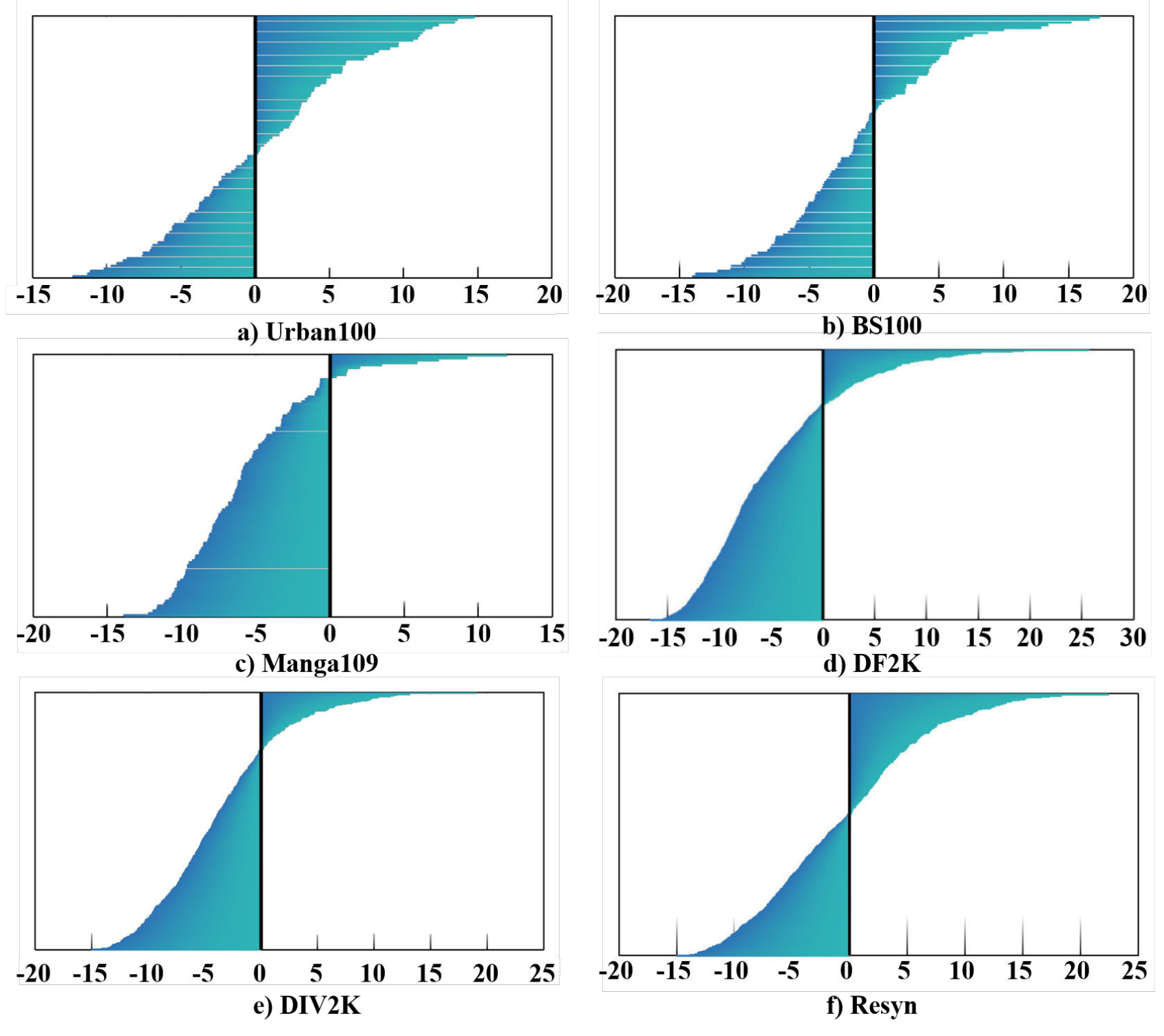}      
    \caption{The complexity distributions {of} different datasets. The {complexity} distribution{s} of the training datasets DIV2K~\cite{timofte2017ntire} and DF2K{~\cite{lim2017enhanced}} have a typical shift, contain{ing} more images of low complexity. 
    Our ReSyn dataset balances the distribution of low and high complexity images {by image filtering based on the newly proposed GLCM image complexity measure}.} 
    \label{fig: Com Dis}
\end{figure}

\subsection{Receptance Weighted Key Value (RWKV)}
The attention mechanism has shown promising performance in both CV and NLP fields. Various operators with linear complexity~\cite{peng2023rwkv,gu2023mamba,qin2023toeplitz} have been explored to optimize the global attention mechanism in recent years. A modified form of linear attention, the Attention Free Transformer (AFT)~\cite{zhai2021attention}, paved the way for the RWKV architecture by using some attention heads equal to the size of the feature dimension and incorporating a set of learned pairwise positional biases. RWKV-v4~\cite{peng2023rwkv} employed exponential decay to model global information efficiently. Vision-RWKV~\cite{duan2024vision} transfers the RWKV-v4 to the vision domain through a Q-shift mechanism and bidirectional attention. RWKV-5/6~\cite{peng2024eagle} further refined the architecture of RWKV-4. RWKV-5 adds matrix-valued attention states, LayerNorm over the attention heads, SiLU attention gating, and improved initialization. It also removes the Sigmoid activation of receptance. RWKV-6 further applies data dependence to the decay schedule and token shift. We modify the RWKV-v4 module of Vision-RWKV, use a depth-wised conv to replace the Q-shift, and also a Cross-Bi-Direction attention to replace the Bi-Direction attention.

\section{ReSyn Dataset} \label{section:dataset}

Due to nd{limited photography techniques and compression techniques}, many images in previous  datasets~\cite{deng2009imagenet,timofte2017ntire} {suffer from} noise, blur{ring}, and other problems. 
However, most {datasets} focus solely on obtaining high-resolution images, {using resolution and Bits Per Pixel (BPP) for image filtering, but lacking sufficient consideration for image complexities.}
This causes a distribution bias {(different image complexities)} between the classic image restoration training datasets and the test datasets. 
We {propose} an image complexity {metric based on} the Gray Level Co-occurrence Matrix (GLCM) analysis method to directly {analyze} this {complexity distribution} bias. 
As shown in Fig.~\ref{fig: Com Dis}, we analyze the GLCM complexity distribution of the classic SR training dataset{s} {(}DIV2K~\cite{timofte2017ntire} {and DF2K~\cite{lim2017enhanced}),} and test datasets {(}Urban100, Manga109 {and BS100)}. 
{It can be seen that the classic SR training datasets and test datasets often have different image complexity distributions.
And the}
test dataset {that often achieves a better PSNR performance, \textit{e.g.,} Manga109, has} a complexity distribution closer to that of the training dataset.
We further analyze the relation{ship} between the GLCM complexity indicator and the restoration performance metric PSNR.
As shown in Fig.~\ref{fig: relation}, GLCM complexity can better predict the {restoration} performance compared to the {BPP} indicator.  
{Moreover}, datasets {with} lower complexity show better restoration performance.

To enhance the performance of existing methods and {incorporate} images generated by the AIGC method, we {construct} the ReSyn dataset, a new large-scale dataset for {both} the Real and Synthetic image ReStoration. 
Our dataset introduces the GLCM complexity indicator as a criterion {for filtering images to} help improve the quality of shuffled images {and achieve a balanced complexity distribution}. 
Examples of the images from our ReSyn dataset are shown in Fig.\ref{fig: Diversity}. The data collection pipeline and the dataset analysis are introduced {in details below}. 


\subsection{DATA COLLECTION}
\noindent\textbf{{Data} Source.}
{Our dataset consists of real and AI-generated images.}
Previous methods~\cite{chen2023activating, zhou2023srformer} improve performance by pretraining on large-scale datasets like ImageNet~\cite{deng2009imagenet}. 
We follow this trend and borrow images from these datasets {to form} the real image part of our dataset. 
The real images are collected from the commonly used large-scale datasets for high-level tasks, including ImageNet~\cite{deng2009imagenet}, COCO2017~\cite{lin2014microsoft}, and SAM~\cite{sam}. The images are shuffled and low-quality images are discarded, only 9K images are retained. 
To introduce the {AI-generated} images into this dataset, we automatically crawl images from Midjourney{,} and {after filtering we} retain 3K {synthetic} images for our dataset. The origins of these image distribution charts and the diversity of their resolutions are illustrated in Fig.~\ref{fig: Diversity}. 
{Different from previous datasets that filter solely based on resolutions, w}e do not completely discard images with resolutions below 2K, {since} many images in the test dataset are below 2K resolution and have a high image complexity. Our dataset encompasses a broad {range} of image resolutions from 0.5K to 4K, and most images have a resolution {larger than} 1K.

\noindent\textbf{Data Selection Criteria.}
The images {used} for image restoration {model training} need to have a high pixel-level quality. 
{To this end, w}e divide the shuffle process into three steps. 
{1) Firstly}, the images of resolution {smaller} than $800\times800$ are discarded, since for super-resolution tasks{,} the images need to be {down-sampled}. This can help remove most low-quality images. 
{2)} Second{ly}, {to remove the {blurry or noisy degraded} images,} 
we follow the blur and noise suppression process proposed in LSDIR~\cite{li2023lsdir}. The {remaining} images are under blur detection {by} the variance of the image Laplacian{,} and flat region detection through the Sobel filter. 
{3)} Third{ly}, all the images are shuffled through the GLCM complexity {metric (detailed below)} to ensure a balanced distribution. 
We {ensure} that the number of images {with} complexity {values} below zero is equal to that above zero. Therefore, we can form a dataset of balanced image complexity distribution. It should be mentioned that images from different sources are filtered individually.

\begin{figure}[ht]
     \centering
     \begin{subfigure}[b]{0.45\textwidth}
         \centering
         \includegraphics[width=0.98\textwidth, height=0.6\textwidth]{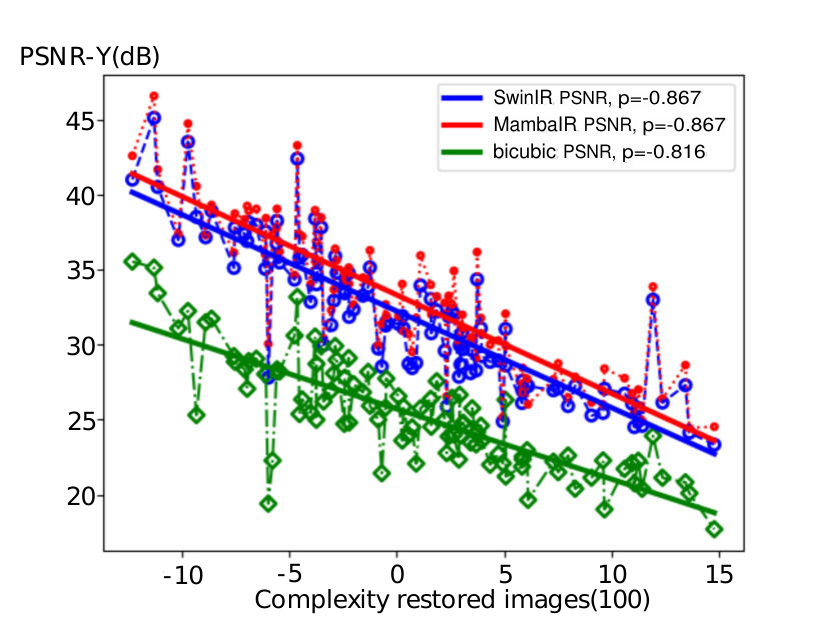}
         \label{fig: complexity}
     \end{subfigure}
     \begin{subfigure}[b]{0.45\textwidth}
         \centering
         \includegraphics[width=0.98\textwidth, height=0.6\textwidth]{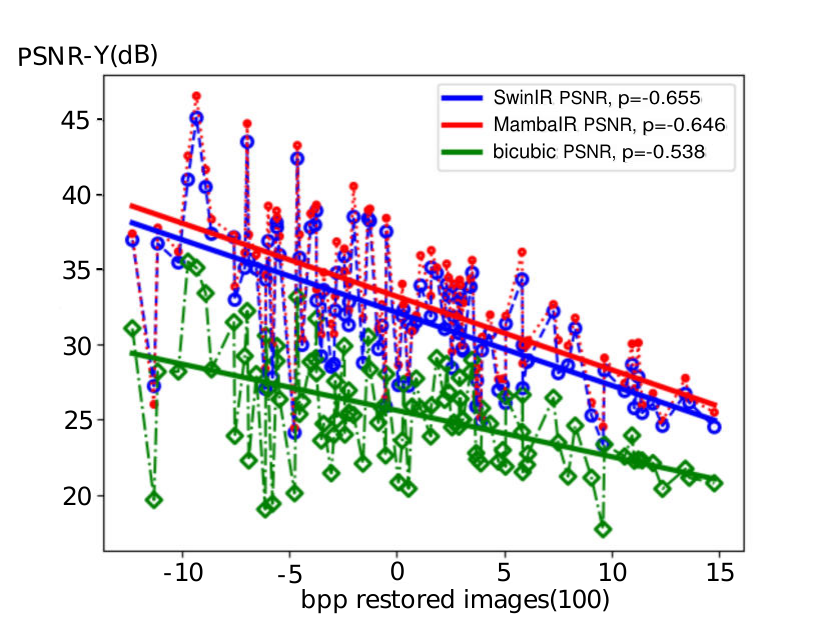}
         \label{fig: bpp}
     \end{subfigure}
        \caption{PSNR ($\times$2 SR on Urban100~\cite{huang2015single}) performance can be predicted by the {proposed GLCM} image complexity and {BPP}~\cite{timofte2017ntire}. We conduct this analysis on MambaIR, SwinIR, and bicubic upsampling restored images.
        }
        \label{fig: relation}
\end{figure}

\begin{figure*}[!t]
\centering
\includegraphics[width=0.8\textwidth]{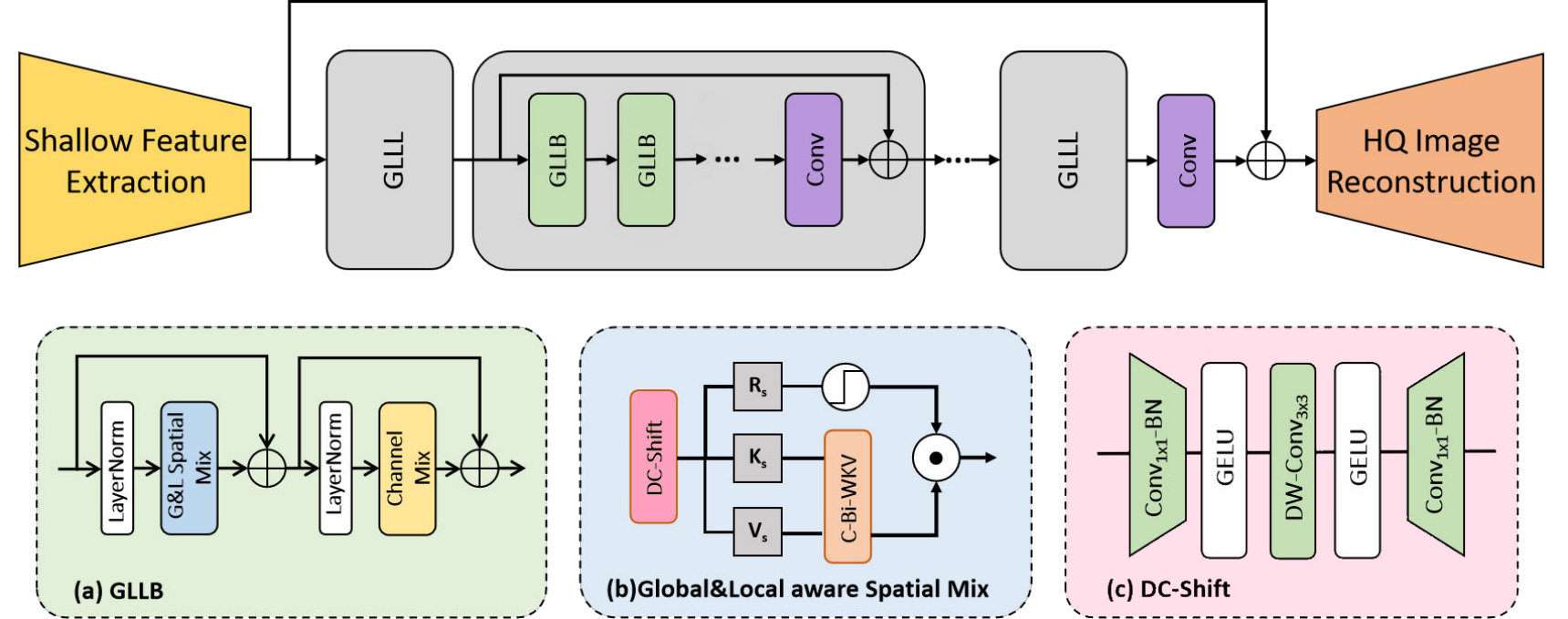}
\caption{\textbf{Framework of our RWKV-IR}, which {consists of three stages:} shallow feature extraction, deep feature {enhancing}, and HQ {image} reconstruction. {For deep feature enhancing, a series of Global$\&$Local Linear attention Layers (GLLL, which is based on RWKV) and a Conv Block are used.} Each GLLL contains several GLLB blocks.}
\label{fig:framework}
\end{figure*}

\noindent\textbf{Image Complexity Analysis.} 
As shown in previous works~\cite{timofte2017ntire}, the PSNR metric of super-resolution images is strongly correlated with the {B}its {P}er {P}ixel (BPP), {an} indicator {for} the quantity of image information. 
nd{{Although} BPP reflects the quality of an image by measuring its color depth, it lacks consideration for the relationships between pixels and cannot adequately measure the texture variation and complexity of an image. 
Therefore,}
this paper further {measures the image complexity} based on the Gray Level Co-occurrence Matrix (GLCM) {and investigates its correlation with PSNR metric}. 
Since human eyes are more sensitive to texture, we utilize the GLCM, which is closely related to the complexity of image texture, to construct an image complexity analysis {metric}. 
We calculate relevant statistical quantities from the GLCM{,} and use Entropy, Energy, and Dissimilarity to build a formula for image complexity analysis as follows: $I_{complexity} =  ENT - ENE + DISS$,
where $ENT$, $ENE$, and $DISS$ represent entropy, energy, and dissimilarity {respectively}, all of which are statistical quantities calculated from GLCM. 
As shown in Fig.~\ref{fig: relation}, we {analyze} the correlation between the PSNR-Y metric and {GLCM-based} image complexity, as well as {BPP}. 
The PSNR {metrics are measured on super-resolution results generated by} two pre-trained models and a direct bicubic upsample for the Urban100 test images{, and} sorted according to {GLCM} complexity and {BPP}. 
Our {GLCM} complexity measure has a stronger Pearson correlation ($\rho=-0.86$) compared to {BPP} ($\rho=0.65$), indicating that the proposed image complexity is a stronger predictor for the PSNR metric of the restored images. Furthermore, the distribution of {GLCM} complexity is symmetric {with respect to} the origin, making it an excellent {metric} for image filtering. \textbf{Post Processing.} For the classic image restoration tasks, {\textit{e.g.,}} super-resolution, the training {requires} paired down-sampl{ed LR} images and GT. We employ the classic bicubic {down-sampling} method\footnote{Simulating MATLAB's anti-aliasing imresize using a Python-based approach, with negligible differences in visual effects.} to {obtain the LR images}. We use the common{ly used} scale factors of $\times2$, $\times3$, and $\times4$.

\noindent\textbf{Partitions.} After the shuffle and the post-processing, {there are} 12K images left, {of which} 9K are real images and 3K are synthetic images. We {then} randomly partition {our ReSyn dataset into a training set of} 10K images, {a} validation {set of} 1K images, and {a} test {set of} 1K images.

\section{Methodology: RWKV-IR} \label{section:method}

The Receptance Weighted Key-Value model (RWKV)~\cite{peng2023rwkv} {is} a linear complexity attention mechanism {that} combines {the advantages of} RNN and {Transformer}. 
{Its linear complexity} enables the utilization of a broader range of pixels for activation{, which is suitable for image restoration tasks}. 
Consequently, we integrate this linear complexity attention mechanism with a classic image restoration model to construct our \textbf{RWKV-IR}. 
The framework of our RWKV-IR is illustrated in Fig.~\ref{fig:framework}, following the widely used structure~\cite{liang2021swinir,guo2024mambair} that consists of three {stages}: shallow feature extraction, deep feature {enhancing}, and high-quality {image} reconstruction{, where the RWKV is mainly incorporated into the second stage}. 
{1) Shallow feature extraction:} Given a low-quality input 
$I_{LQ} \in \mathbb{R}^{H \times W \times 3}$, a $3 \times 3$ convolution layer first extracts the shallow feature $F_{S} \in \mathbb{R}^{H \times W \times C}$, where $H$ and $W$ represent the height and width, and $C$ is the number of channels in the {shallow} feature. 
{2) Deep feature enhancing:} Subsequently, a series of {Global$\&$Local Linear attention Layers} {(GLLL, which is based on RWKV)} and a $3 \times 3$ Convolution Block perform deep feature extraction. Each GLLL layer contains several GLLB blocks. 
Afterward{s}, a global residual connection fuses the shallow feature $F_{S}$ and the deep feature $F_{D}$ {into a hybrid feature} $F_{H} = F_{S} + F_{D}${, which is then} input {into the} high-quality {image} reconstruction {module}. 
{3) High-quality image reconstruction:}
Finally, the high-quality reconstruction {module outputs} a restored image $I_{RE} \in \mathbb{R}^{(H \times s) \times (W \times s) \times 3}$, where $s$ is the scale factor used for the super-resolution task.

\noindent\textbf{Global$\&$Local Linear attention Block (GLLB).} 
Transformer-based restoration networks~\cite{liang2021swinir,chen2023activating} typically design the core block for restoration following a $norm \rightarrow Attention \rightarrow norm \rightarrow MLP$ {workflow}. 
The {$Attention$} module is designed {to model} global dependency, but due to heavy computational complexity, only local window attention is used. Therefore, replacing the local attention with a linear complexity attention mechanism can reduce the computational overhead {while increasing} the window size, {to better} model global dependencies. 
Therefore, we {aim} to introduce the Receptance Weighted Key Value {(RWKV)} module to enhance the image restoration effect{s}. 
However, simply replacing the $Attention$ module with the Spatial Mix from RWKV, and {replacing} the $MLP$ module with the Channel Mix from RWKV yields sub-optimal results {(as shown in {Tab.~\ref{tab:shift ablation},  \ref{tab:FFN ablation}})}. 
We find that the linear complexity attention module from RWKV can model global dependencies well, but 
{its utilization of local information is insufficient.}
{Through} further experiments, we find that this is caused by the {Q-}shift and the Bi-direction WKV {in the original RWKV,} which are transferred from the NLP tasks and not suitable for the low-level {vision} tasks. 
Therefore, we replace the {Q-}shift with a {newly proposed} {\textbf{Depth-wise Convolution shift {(DC-shift)}}} to achieve better visual representation and easier optimization~\cite{yuan2021incorporating,zhao2021battle,chen2023activating}{. 
Moreover, we propose} a {\textbf{Cross-Bi-WKV module} that integrates two Bi-WKV modules with different scanning orders,} {instead of the original Bi-WKV,} to {balance the attention for horizontal and vertical directions and} improve the model performance.

\begin{wrapfigure}{r}{4cm} 
\includegraphics[width=4cm]{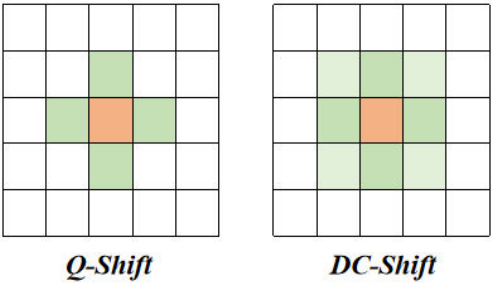}
\caption{Different shift methods. {The Q-shift is a simple channel replacement operation using four {neighboring} pixels, while our DC-shift is a depth-wise conv leveraging {the} surrounding pixels {in a $k \times k$ neighborhood}.}}\label{fig:shift}
\end{wrapfigure}

As shown in Fig.~\ref{fig:framework}(a), {we propose a \textbf{Global$\&$Local Linear attention Block (GLLB)}, which} follow{s} a $norm \rightarrow Conv \rightarrow Attention \rightarrow norm \rightarrow {Channel\text{-}Mix}$ {workflow, with two residual connections}. 
{Given} an intermediate feature $F_{i}$, where $i$ represents the $i$-th GLLB {block}. 
A LayerNorm module {is} followed by {a} linear complexity \textbf{Global{$\&$Local}-Aware Spatial Mix} module{, that models both} long-term dependencies {and local dependencies}: $F_{g,i} = {GL}Spatial\text{-}Mix(LN(F_{i}))$. 
{The local-aware characteristics are achieved by replacing the Q-Shift in the original Spatial Mix with our Depth-wise Convolution shift (DC-Shift, detailed below).}
Furthermore, the Channel Mix module replacing the $MLP$ is used for stabilizing the training process and avoiding channel oblivion: $F_{i+1} = F_{g,i}*\beta + LN({Channel\text{-}Mix}(F_{g,i}))$.
\\
\noindent\textbf{{DC-Shift}.} 
{To emulate the memory mechanism of RNNs, the original RWKV propose{s} a {token shift} mechanism.} 
Consider an input feature $X\in \mathbb{R}^{T\times C}$ (where $T=H\times W$), it is first shifted{,} and {then} fed into three linear layers to obtain the matrices $R_s,K_s,V_s\in\mathbb{R}^{T\times C}$:
$R_s={Shift}_{R}(X)W_R, K_s={Shift}_{K}(X)W_K, V_s={Shift}_{V}(X)W_V.$
{Then}, $K_s$ and $V_s$ are {used} to calculate the global attention 
$wkv\in\mathbb{R}^{T\times C}$ by a linear complexity bidirectional attention mechanism, and multiplied with $\sigma(R_s)$ which controls the output $O_s$ probability. {But the original {implementation of $Shift$ is a} Q-shift {operation, which} simply {combines the features from the top, left, down, and right neighboring pixels, each using $C/4$ channels, to replace the feature of the center pixel, formulated as follows}: }
\begin{equation}
\begin{aligned}
\label{eq:Quad_sf}
    {Q\text{-}Shift}_{(*)}(X)=X+(1+{\mu}_{(*)})X',
    where X'[h,w]=\\Concat(X[h-1,w,0:C/4],X[h+1,w,C/4:C/2],\\
     X[h,w-1,C/2:3C/4],X[h,w+1,3C/4:C]).\\
\end{aligned}
\end{equation}

{{The Q-Shift is not suitable for image restoration due to two reasons: 1) In image restoration tasks,} the number of channels in the features is relatively small compared to NLP {tasks;
and 2)}
The {simple} feature substitution {in Q-Shift does not consider the similarity between local pixels, making it} not suitable for 
{image restoration tasks that rely on local similarity.}
Therefore, we {propose} a \textbf{Depth-wise Convolution Shift {(DC-Shift)}} shown in Fig.~\ref{fig:shift} to replace the Q-shift{, which} help{s enhance} the model performance {by modeling the relationships in local receptive fields}.} 
{As} shown in Fig.~\ref{fig:framework}(c), the Depth-wise Convolution shift consists of two $1\times1$ Convolution Layers and one {$k\times k$} Depth-wise Convolution Layer. By using this structure, we can reduce the number of parameters compared to using the classic channel convolution module and also {compensate} for the lack of local features. 
The {calculation} process of this {DC-Shift} module is {formulated} as: $F_{l,i} = Conv_{1\times 1}(GeLU(DW\text{-}Conv_{k\times k}(GeLU(Conv_{1\times 1}(X)))))$, where $k$ is the kernel size of the depth-wise convolution.
{By combining the Depth-wise Convolution Shift with Bi-directional attention, as shown in Fig.~\ref{fig:framework}(b), we achieve Global$\&$Local-Aware Spatial Mix.}

\begin{table*}[!t]
\centering
\caption{Quantitative comparison on \underline{\textbf{classic image super-resolution}} with state-of-the-art methods on 10K iters training. The best and the second best results are in \best{red} and \second{blue}.}
\label{tab:classicSR}
\renewcommand{\arraystretch}{0.9}
\renewcommand{\arraystretch}{1.0}
\setlength\tabcolsep{3.0pt}
\resizebox{1.\linewidth}{!}{
\begin{tabular}{@{}l|c|c|cc|cc|cc|cc|cc|cc@{}}
\hline 
 & & & \multicolumn{2}{c|}{\textbf{Set5}} &
  \multicolumn{2}{c|}{\textbf{Set14}} &
  \multicolumn{2}{c|}{\textbf{BSDS100}} &
  \multicolumn{2}{c|}{\textbf{Urban100}} &
  \multicolumn{2}{c|}{\textbf{Manga109}} &
  \multicolumn{2}{c}{\textbf{ReSyn}}\\
\multirow{-2}{*}{ Method} & \multirow{-2}{*}{scale} & \multirow{-2}{*}{dataset} & PSNR  & SSIM   & PSNR  & SSIM   & PSNR  & SSIM   & PSNR  & SSIM   & PSNR  & SSIM  & PSNR  & SSIM \\ \hline
HAN~\cite{niu2020single}    & $\times 2$ & DF2K &   38.26   &  0.9611    &   34.01  &  0.9205   &  32.36   &   0.9008   &   33.09   &   0.9366   &  39.56   &  0.9788    &  35.22   &0.9311\\
SwinIR~\cite{liang2021swinir} & $\times 2$ & DF2K & 38.25 & 0.9616 & 34.04 & 0.9215 & 32.39 & 0.9024 & 33.06 & 0.9365 & 39.54 & 0.9790 & 35.24 & 0.9313 \\
SRFormer~\cite{zhou2023srformer} & $\times 2$ & DF2K  & 34.80 & 0.9301 & 30.74 & 0.8495 & 29.33 & 0.8113 & 29.12 & 0.8712 & 34.68 & 0.9510 & 32.30 & 0.8763\\
MambaIR~\cite{guo2024mambair} & $\times 2$ & DF2K  & \second{38.34} & 0.9617 & {34.42} & \second{0.9246} & 32.45 & 0.9032 & 33.55 & 0.9401 & \best{39.78} & \best{0.9796} & 35.40 & 0.9325\\
\best{RWKV-IR (Ours)}  & $\times 2$ & DF2K &  \best{38.38}  & 0.9618   &  34.42 & 0.9246  &  {32.46} &\best{0.9034} & \best{33.59} &  \best{0.9404} &  39.80 &  0.9797 &  {35.44} & {0.9327} \\
\hdashline
HAN~\cite{niu2020single}    & $\times 2$ & ReSyn &   38.08   &  0.9002    &  33.88   &   0.9198  & 32.30    &  0.9011    &  33.10    & 0.9352     &  39.15   &  0.9733    &  35.46   &0.9328\\
SwinIR~\cite{liang2021swinir} & $\times 2$ & ReSyn &  38.18 & 0.9616 & 34.03 & 0.9213 & 32.39 & 0.9026 & 33.12 & 0.9370 & 39.30 & 0.9791 & 35.52 & 0.9331 \\
SRFormer~\cite{zhou2023srformer} & $\times 2$ & ReSyn & 38.17 & \best{0.9621} & 34.12 & 0.9224 & 32.41 & 0.9031 & 33.39 & 0.9396 & 39.50 & 0.9797 & 35.62 & 0.9340\\
MambaIR~\cite{guo2024mambair} & $\times 2$ & ReSyn &  38.26 &  0.9616  &  \best{34.43}  & \best{0.9247}  & \second{32.46} &  \second{0.9034}  & 33.54 & 0.9401 &  39.72 & 0.9794  & \second{35.64}    &  \second{0.9350}\\
\best{RWKV-IR (Ours)}  & $\times 2$ & ReSyn & 38.28 & \second{0.9618} & \second{34.42} & 0.9245 & \best{32.47} & 0.9032 & \second{33.58} & \second{0.9404} & \second{39.76} & \second{0.9796} & \best{35.68} & \best{0.9352}  \\
\hline
HAN~\cite{niu2020single} & $\times$3 & DF2K &  34.77    &   0.9300   & 30.60    &  0.8466   &  29.28   &  0.8110    &  29.03    &  0.8701    &  32.56   &  0.9451    &  32.25   &0.8744\\
SwinIR~\cite{liang2021swinir} & $\times$3 & DF2K &34.80 & 0.9301 & 30.74 & 0.8495 & 29.33 & 0.8113 & 29.12 & 0.8712 & 34.65&0.9501 & 32.30 & 0.8763\\
SRformer~\cite{zhou2023srformer} & $\times$3 & DF2K  & 34.62 & 0.9306 & 30.73 & 0.8501 & \second{29.39} & \best{0.8132} & \best{29.53} & \best{0.8784} & 34.92 & \best{0.9524} & 32.38 & 0.8779\\
MambaIR~\cite{guo2024mambair} & $\times$3 & DF2K  & \second{34.86} & \second{0.9307} & \second{30.76} & \second{0.8505} & 29.39 & 0.8123 & 29.42 & 0.8758 & \second{34.92} & 0.9519 & 32.42 & 0.8780\\
\best{RWKV-IR (Ours)}  & $\times 3$ & DF2K & \best{34.88}  &  \best{0.9308}   & \best{30.77}   & \best{0.8507}   & \best{29.40}  & \second{0.8124} &  29.43    &  0.8757    &  \best{34.95}   &  \second{0.9520}    & 32.43    & 0.8782\\
\hdashline
HAN~\cite{niu2020single} & $\times$3 & ReSyn &  34.49    &0.9278 & 30.36    & 0.8427    & 29.20    &  0.8110   &   28.73   &0.8711            &   34.40  &  0.9500    &   32.38  &0.8772\\
SwinIR~\cite{liang2021swinir} & $\times$3 & ReSyn &  34.68 & 0.9296 & 30.71 & 0.8487 & 29.30 & 0.8113 & 29.20 & 0.8723 & 34.44 & 0.9505 & 32.41 & 0.8776 \\
SRformer~\cite{zhou2023srformer} & $\times$3 & ReSyn  & 34.48 & 0.9298 & 30.72 & 0.8488 & 29.34 & 0.8127 & \second{29.46} & \second{0.8771} & 34.53 & 0.9512 & 32.47 & 0.8786 \\
MambaIR~\cite{guo2024mambair} & $\times$3 & ReSyn  &  34.74    &  0.9298    &  30.74  & 0.8492    & 29.35  &  0.8122    &   29.42   & 0.8766 & 34.53 & 0.9511   & \second{32.49} & \second{0.8803} \\
\best{RWKV-IR (Ours)}  & $\times 3$ & ReSyn & 34.77     &  0.9299    & 30.73    & 0.8491    &  29.34   & 0.8123     &   29.45   & 0.8768     & 34.55    &  0.9513    &  \best{32.50}   & \best{0.8805}\\
\hline
HAN~\cite{niu2020single}      & $\times 4$  & DF2K &    32.51  &    0.9001  &  28.85   &  0.7856   & 27.75    &  0.7440    &  26.88  &0.8155  & 31.72     & 0.9015    &  30.60   & 0.8321    \\
SwinIR~\cite{liang2021swinir}    & $\times 4$ & DF2K & \second{32.74} & \best{0.9020} & 29.02 & 0.7920 & 27.83 & 0.7457 & 27.12 & 0.8162 & 31.75 & 0.9223 & 30.67 & 0.8335\\
SRFormer~\cite{zhou2023srformer} & $\times 4$ & DF2K  & 32.68 & 0.9010 & 29.03 & 0.7917 & 27.82 & 0.7458 & 27.38 & 0.8218 & \second{31.86} & \best{0.9234} & 30.72 & 0.8346\\
MambaIR~\cite{guo2024mambair}   & $\times 4$  & DF2K  & 32.73 & 0.9015 & \best{29.05} & \second{0.7923} & \best{27.87} & \best{0.7465} & 27.20 & 0.8176 & 31.83 & 0.9228 & 30.73 & 0.8345\\
\best{RWKV-IR (Ours)}  & $\times 4$ & DF2K &  \best{32.74}   &   \second{0.9016}   & \second{29.04}    & \best{0.7924}    &  \second{27.86}   & \second{0.7466}     &  27.21    & 0.8177     & \best{31.87}    & \second{0.9233} &30.74 &0.8350  \\
\hdashline
HAN~\cite{niu2020single}      & $\times 4$  & ReSyn &  32.29    &  0.8991   &   28.59  &  0.7878   & 27.64    &  0.7420    &  26.58    & 0.8155     &  31.55   & 0.9201    &  30.67   &0.8321 \\
SwinIR~\cite{liang2021swinir}    & $\times 4$ & ReSyn &  32.63 & 0.9009 & 28.98 & 0.7908 & 27.81 & 0.7449 & 27.16 & 0.8158 & 31.60 & 0.9220 & 30.75 & 0.8342 \\
SRFormer~\cite{zhou2023srformer} & $\times 4$ & ReSyn & 32.61  & 0.9007     &  28.98 & 0.7909    &  27.82   &  0.7452  &  \best{27.42}    &  \best{0.8220}  &  31.72  &  0.9225 &  30.79   & 0.8350 \\
MambaIR~\cite{guo2024mambair}   & $\times 4$  & ReSyn & 32.65      &  0.9009  &  29.00   & 0.7910 & 27.85  &  0.7458    &   {27.21}   &  0.8181 &  31.79   &  0.9224 & \second{30.81} & \second{0.8350}\\
\best{RWKV-IR (Ours)}  & $\times 4$ & ReSyn &  32.66    & 0.9011     & 29.02     & 0.7911    &  27.86   &  0.7459    & \second{27.24} &  \second{0.8183}    & 31.81    & 0.9226     & \best{30.83}    &\best{0.8351}\\
\hline 
\end{tabular}%
}
\end{table*}

\noindent\textbf{Cross-Bi-WKV module.} 
The core idea of the {Vision-}RWKV is the linear complexity Bi-directional attention {(Bi-WKV)}. {Its calculation result for the $t$-th pixel is {below}:} 
\begin{align}
    \label{eq:wkv-recur}
    wkv_t = Bi\text{-}WKV(K,V)_t =  \\\frac{\sum_{i=0,i\neq t}^{T-1}e^{-(\lvert t-i \rvert -1)/T\dot w+k_i}v_i + e^{u + k_t}v_t}{\sum_{i=0,i\neq t}^{T-1}e^{-(\lvert t-i \rvert -1)/T\dot w+k_i}v_i + e^{u + k_t}}, 
\end{align}

\begin{figure}[t]
  \centering
  \includegraphics[width=0.45\textwidth]{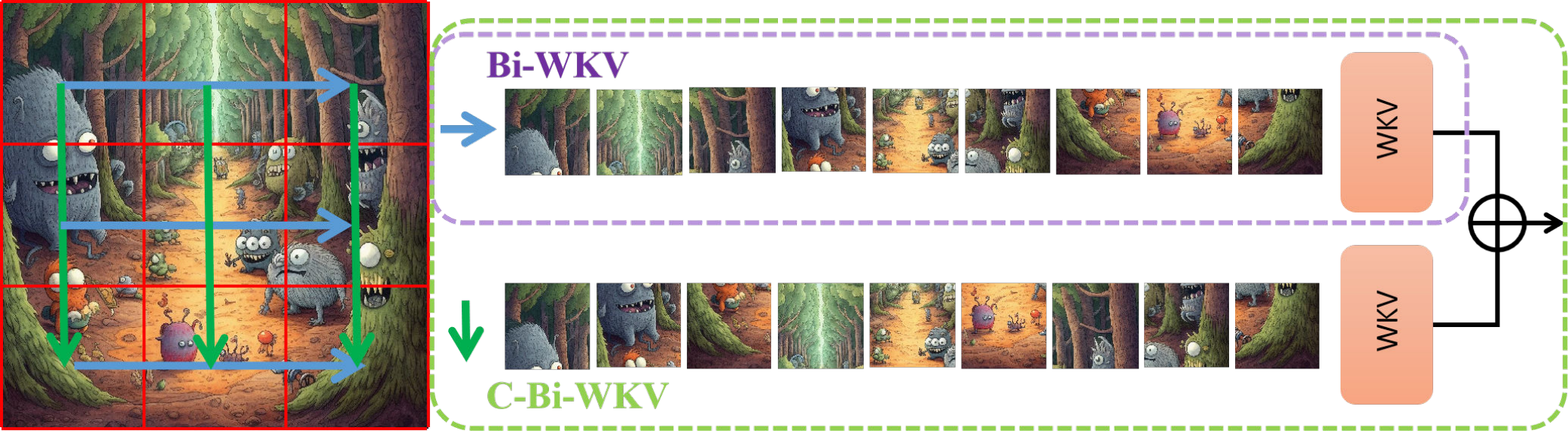}
  \caption{{Illustration of} Cross-Bi-WKV, {which consists of two cross scanning Bi-WKV {modules}.}}
  \label{fig:cwkv}
\end{figure}

{
{where the upper limit for the current pixel $t$ (the $t-th$ pixel after {flattening the 2D pixels into a 1D sequence using} the horizontal scan{ing order}) is set}
to $T-1$ (the last pixel), to ensure that all pixels are mutually visible in the calculation of each {other's} result. 
{In this formula, $(\lvert t-i \rvert-1)/T$ is used as the position embedding, which is unbalanced for horizontal and vertical directions, \textit{i.e.,} the position embedding differences between the left and right neighboring pixels are much smaller than that between the up and down neighboring pixels, and after applying the negative sign and exponential calculation, this leads to more attention to the horizontal direction than the vertical direction.}} 
{Therefore,} we {propose} a Cross-Bi-directional {WKV module}, {which {combines} a horizontal direction Bi-WKV and a vertical direction Bi-WKV to {achieve a balanced} attention to horizontal and vertical pixels.} 
{The two Bi-WKV modules use different scanning orders when flattening the pixels into a 1D sequence, one using the horizontal scanning order, and the other using the vertical scanning order,}
nd{as shown in Fig.~\ref{fig:cwkv}}. {And w}e use the average {output} of the two Bi-WKV {modules} to form the {final} output feature. 
After the DC-Shift, {our Cross-Bi-directional} attention  mechanism {with} a linear complexity is {formulated as follows} and {outputs} a global attention result $wkv\in\mathbb{R}^{T\times C}$: $wkv=C\text{-}Bi\text{-}WKV(K,V)=Bi\text{-}WKV_{horizontal}(K_h,V_h)+Bi\text{-}WKV_{vertical}(K_v,V_v)$. {The {result} $wkv$ {is} then multipl{ied} with $\sigma(R)$ to {obtain} the output $O_s$ probability: $O_s=(\sigma(R_s)\bigodot wkv)W_O$.}{
{As the} RWKV {model continues to} iterat{e} and update, we believe that subsequent versions will bring greater improvements.}

\section{Experiments} \label{section:exp}
Few previous methods have focused on the impact of training criteria on the performance comparison of restoration models. Many approaches~\cite{liang2021swinir,zhou2023srformer,guo2024mambair} have instead opted to extend training time continuously to improve model performance. This can lead to unfair comparisons in subsequent evaluations. Therefore, we construct a comprehensive training benchmark from two perspectives: measuring the model's convergence ability and its restoration capability. In this section, we present the experimental benchmark and the training details of our proposed linear-complexity attention-based image restoration model. We then conduct a benchmark study comparing our method with other models. \textit{Due to space limitations, only the results of classical SR tasks are shown in {the main paper}}. Please refer to {the \textbf{Appendix} for {the experiments of} other IR tasks {(light-weight SR, image denoising, JPEG artifacts reduction). Source codes are provided in the supplementary material}}.

\subsection{Experimental Settings}
\textbf{Experimental Benchmark.} We propose a unified training benchmark for different kinds of image restoration tasks. \textit{1) Super Resolution:} The commonly conducted SR tasks are lightweight SR and classical SR. For all compared methods, we set the same batch size and the same number of iterations for training. To show models' convergence and restoration abilities, we compared different models in two number levels of training iterations. The batch size for lightweight SR model training is 64, and the training iterations are 50K and 500K. The batch size for classical SR model training is 32, and the training iterations are 100K and 500K. \textit{2) Image Denoising:} For Gaussian color denoising and gray-scale denoising tasks, the training batch size and the training iterations are set to 16 and 100K (500k for long training), respectively. \textit{3) JPEG Compression Artifact Reduction:} the batch size and training iterations are set to 16 and 100K (500k for high number level), respectively. \textit{For more details about the model settings, please refer to the {\textbf{Appendix}.}}
\\
\textbf{Training Details.} We conduct super-resolution (SR) training experiments on three datasets: DIV2K~\cite{timofte2017ntire}, DF2K~\cite{lim2017enhanced}, and {our} ReSyn. For lightweight SR, models are trained separately on the DIV2K and ReSyn datasets. For classic SR, models are trained separately on the DF2K and ReSyn datasets. In the training of image denoising models, we compare performance on the DFWB RGB dataset (a combined dataset of DIV2K~\cite{wang2023selective}, Flickr2K~\cite{lim2017enhanced}, BSD500~\cite{arbelaez2010contour}, and WED~\cite{ma2016waterloo}) and our ReSyn dataset. 
During training, we crop the compressed images into $64\times64$ patches for image SR. We do not use pre-trained weights from the $\times2$ model to initialize those of $\times3$ and $\times4$ but train the $\times3$ and $\times4$ models from scratch. For the denoising task, we crop the original images into $128\times128$ patches. We employ the Adam optimizer for training our RWKV-IR with ${\beta}_1 = 0.9$ and ${\beta}_2 = 0.999$. The initial learning rate is set at $2 \times 10^{-4}$ and is decreased during training using the multi-step scheduler. Our models are trained with 8 NVIDIA V100 GPUs. Except for the batch size and training iter{ation}s for the compared methods, all other settings remain consistent with their official codes.

\subsection{Comparison on Image Super-Resolution}
\textbf{Classic Image Super-Resolution.} Table~\ref{tab:classicSR} presents quantitative comparisons between RWKV-IR and state-of-the-art methods (HAN~\cite{niu2020single}, SwinIR~\cite{liang2021swinir}, SRFormer~\cite{zhou2023srformer}, and MambaIR~\cite{guo2024mambair}) on 100K training iters, which can show the convergence capability of models. 
Our method achieves optimal results on almost all five datasets for all scale factors. As shown, our RWKV-based baseline outperforms SwinIR by 0.08dB on Urban100 for x4 scale and MambaIR by 0.03dB, demonstrating the image restoration capability and quick convergence ability of our RWKV-IR. Furthermore, training our newly constructed ReSyn dataset also achieves decent performance, despite the overall quality of the data sources not being high. This also proves the feasibility of the image complexity-based dataset construction method. \textit{The comparisons on long training iterations {and other IR tasks} could be found in the \textbf{Appendix}.}

\subsection{Ablation Study}
In this section, we conduct {ablation studies} to explore the effects of different designs of the core GLLB on the test dataset Urban100. These experiments can demonstrate the issues that need to be considered when applying the linear attention mechanism RWKV to image restoration models. They also provide an intuitive reflection of the problems mentioned earlier, offering insights for subsequent application research. To reduce training costs, all the models are lightweight models trained on the DIV2K dataset. 
{The ablation studies results in} Tab.~\ref{tab:pos ablation},~\ref{tab:wkv ablation},~\ref{tab:shift ablation} and~\ref{tab:FFN ablation} indicate that:  
{\textbf{\textit{(1)}}} The original Q-Shift from the RWKV hinder{s} the performance of the model on the image restoration task{,} since {its simple feature substitution does not capture local similarity.} 
{\textbf{\textit{(2)}} We propose} the Depth-wise Convolution {Shift (DC-Shift)} to replace the Q-shift, {which models relationships in local receptive fields, helping to enhance the restoration capabilities, thereby} obtain{ing} a {PSNR} improvement of {0.87dB}. 
{\textbf{\textit{(3)}}} Since the Bi-WKV {pays unbalance attention to horizontal and vertical directions, our} Cross Bi-WKV module {that combines two Bi-WKV modules with different scanning orders} further improve{s} the performance of the {image restoration} model{, with a PSNR improvement of 0.75dB}. 
{\textbf{\textit{(4)}}} The MLP module is not suitable for image restoration, a Channel Attention {Block (CAB)} or Channel Mix process can further improve the model performance. These experiments show the Cross-Bi-WKV module and the Depth-wise Conv Shift can help improve the performance. With the continuous iteration and upgrades of RWKV, we believe that subsequent versions of RWKV will bring genuine global attention, significantly enhancing the IR {capabilities}.

\begin{table}[t]
\scriptsize
\centering
\renewcommand{\arraystretch}{1.5}
\setlength{\tabcolsep}{2pt}
\resizebox{1\linewidth}{!}{
\begin{tabular}{c|c|c|c|c|c}
\hline
{DC-Shift} Position & Before SM & Between SM and CM & Behind CM & Parallel & \textbf{Replace {Q}S} \\ \hline
PSNR {$\uparrow$}         & 32.41  &  32.54   & 32.51  & 32.65 & \best{32.95}    \\ \hline 
\end{tabular}
}
\caption{Ablation study of the insert position of Conv. SM, CM, and TS individually present Spatial Mix, Channel Mix, and Q shift.}
\label{tab:pos ablation}
\end{table}

\begin{table}[t]
\centering
\renewcommand{\arraystretch}{1.0}
\setlength{\tabcolsep}{15pt}
\resizebox{0.8\linewidth}{!}{
\begin{tabular}{c|c|c}
\hline
{WKV S}etting & Bi-{WKV} & {\textbf{Cross-Bi-WKV}} \\ \hline
PSNR {$\uparrow$}        & 32.20  & \best{32.95} \\ \hline 
\end{tabular}
}
\caption{ The study of the different settings of WKV scanning methods. Ours settings of are in \textbf{Bold}}
\label{tab:wkv ablation}
\end{table}


\begin{table}[t]
\centering
\renewcommand{\arraystretch}{1.0}
\setlength{\tabcolsep}{2pt}
\resizebox{1\linewidth}{!}{
\begin{tabular}{c|c|c|c|c}
\hline
Shift Method & Q-Shift (p=1) & Q-Shift (p=0, w.o. shift) & \textbf{DC-Shift (ks=3)} & DC-Shift (ks=5) \\ \hline
PSNR {$\uparrow$}        &   32.08    &  32.69   & \best{32.95}   &  32.84   \\ \hline
\end{tabular}
}
\caption{Ablation study of Shift settings ($p$ in Q-Shift is the distance from neighboring pixels to the center; $ks$ in DC-Shift is the kernel size).}
\label{tab:shift ablation}
\end{table}
\begin{table}[t]
\centering
\renewcommand{\arraystretch}{1.0}
\setlength{\tabcolsep}{15pt}
\resizebox{0.8\linewidth}{!}{
\begin{tabular}{c|c|c|c}
\hline
FFN  & MLP & CAB & \textbf{Channel Mix} \\ \hline
PSNR {$\uparrow$} & 32.32 & 32.67 & \best{32.95}   \\ \hline
\end{tabular}
}
\caption{The study of FFN modules (CAB means Channel Attention Block). {Ours settings of are in \textbf{Bold}. The best scores are in \best{red}.}}
\label{tab:FFN ablation}
\end{table}

\section{Conclusion} \label{section:con}

In this paper, we review the task of image restoration. We utilize the {Gray Level Co-occurrence Matrix} to construct an image complexity metric, which demonstrates the bias between {complexity distributions of} classic training datasets and test datasets. Based on this metric, we construct {ReSyn,} a new image restoration dataset that includes both real and generated images {with balanced complexity}. Additionally, we develop a novel benchmark for comparing image restoration models, focusing on two aspects: the convergence speed and the restoration capability. Moreover, from the perspective of linear attention mechanisms, we {propose a novel RWKV-IR model, which} integrate{s} the RWKV into image restoration models, constructing a linear attention-based image restoration model. This inspires further exploration and enhancement of the effective receptive field of the model. {Extensive} experiments demonstrate the effectiveness of our proposed benchmark and {RWKV-IR} model.

{
    \small
    \bibliographystyle{ieeenat_fullname}
    \bibliography{main}
}

\clearpage
\renewcommand{\thefigure}{A\arabic{figure}}
\setcounter{figure}{0}
\renewcommand{\thetable}{A\arabic{table}}
\setcounter{table}{0}
\renewcommand{\thesection}{A\arabic{section}}
\setcounter{section}{0}
\maketitlesupplementary

\section{More details of Our ReSyn Dataset}
\label{apsec: dataset details}

\begin{figure*}[!h]
\centering
\includegraphics[width=0.80\textwidth]{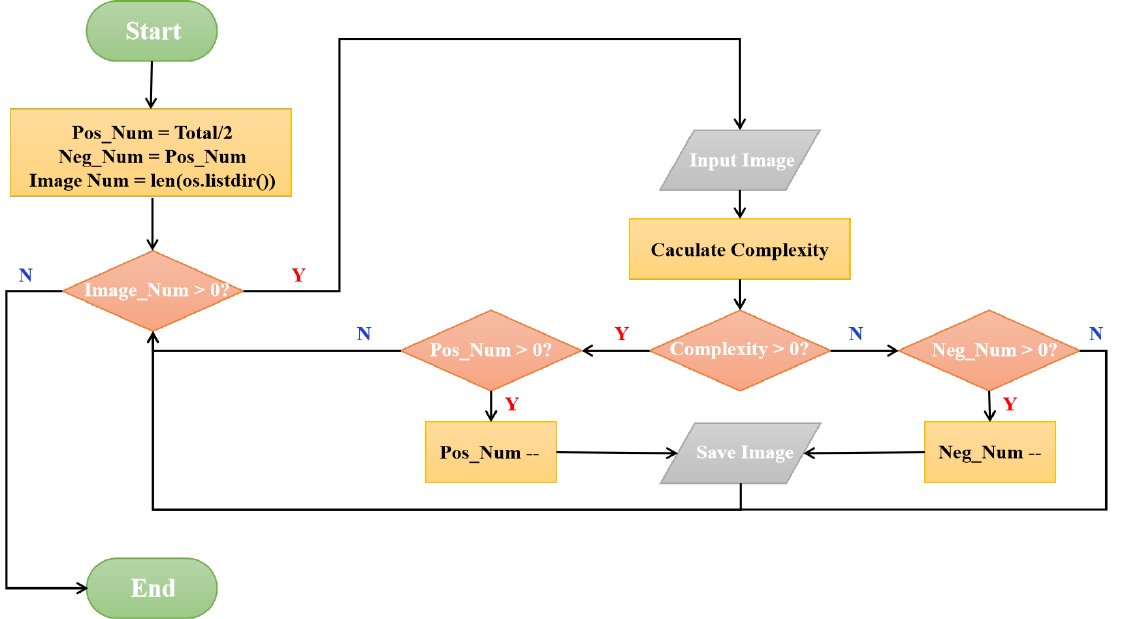}
\caption{The final {filtering} procedure based on {GLCM} image complexity of the ReSyn dataset.}
\label{fig:procedure}
\vspace{0mm}
\end{figure*}

In this section, we further {provide} details on the construction of {our ReSyn} dataset. The process of final filtering based on our {GLCM} image complexity metric is shown in Fig.~\ref{fig:procedure}.

Our ReSyn dataset {contains} images from four sources{,} including ImageNet~\cite{deng2009imagenet}, COCO2017~\cite{lin2014microsoft}, SAM~\cite{sam}, and MidJourney. For ImageNet, we first remove images with a resolution of less than 800$\times$800. Since the details in the images from ImageNet are not rich and relatively blurry{, t}hen the images of blur{ring} and noise degradation are filtered. Finally, through image complexity filtering, we choose 1,200 images from the ImageNet. Half of the images have a complexity greater than zero.

For COCO2017~\cite{lin2014microsoft}, almost all of the images are medium-resolution images, we follow the method of ImageNet filtering and {include} 1,200 images from COCO2017.

For SAM, most images have a resolution of over 2K. For privacy purposes, many of the images in the dataset containing faces and sensitive information are mosaiced. Therefore, we manually remove the images that include the mosaics, leaving only the clear images. After {this step}, we {use} the same {filtering method as} ImageNet and {include} 6,000 images from SAM.

For MidJourney, we first crawled more than 30,000 high-quality images from the web. And then {after filtering,} 3,600 images {are left} to form the dataset.

\section{Model details}
\label{apsec: model details}

{In} Tab.~\ref{tab:model details}, {we provide} the model setting details for different image restoration task{s}, which could serve as a reference for model construction. It should be noted that the number of embed{ding} channels in RWKV-based models must be an integer {that is a} multiple of 16.

\begin{table*}[!ht]
\centering
\caption{The model setting details for different image restoration task{s}.}
\label{tab:model details}
\vspace{1mm}
\renewcommand{\arraystretch}{1.0}
\setlength{\tabcolsep}{16pt}
\resizebox{1\linewidth}{!}{
\begin{tabular}{c|c|c|c|c}
\hline
               & Classic SR        & Light-Weight SR & Denoising         & JPEG              \\ \hline
Embed channel  & 192               & 48              & 192               & 192               \\ \hline
Image size     & 64                & 64              & 128               & 128               \\ \hline
Blocks setting & {[}6,6,6,6,6,6{]} & {[}6,6,6,6{]}   & {[}6,6,6,6,6,6{]} & {[}6,6,6,6,6,6{]} \\ \hline
WKV setting    & Cross WKV         & Layer Cross WKV & Cross WKV         & Cross WKV         \\ \hline
\end{tabular}
}
\end{table*}

\section{More Image Restoration Experiments} 
\label{apsec: model results}

In this section we supplement quantitative comparisons on other image restoration tasks, including {1)} light-weight SR, {2)} image denoising, and {3)} JPEG artifacts reduction. These experiments show the generality of our model.

\subsection{Classical Image Super-Resolution} 
~\label{apsec: class}

In Tab.~\ref{tab:classicSR_500K}, we compare RWKV-IR with other methods on 500K training iter{ation}s. Our newly proposed image restoration models also have a good performance once the training iter{ation} is long. It also has a linear computational complexity, which makes the model save the computational overhead and is more conducive to the scaling of the model.
\begin{table*}[!t]
\centering
\caption{Quantitative comparison on \underline{\textbf{lightweight image super-resolution}} with state-of-the-art methods on 50K training iterations. The best {scores are} in \best{red}.}
\label{tab:lightSR}
\vspace{1mm}
\renewcommand{\arraystretch}{1.0}
\setlength{\tabcolsep}{6pt}
\resizebox{1\linewidth}{!}{
\begin{tabular}{@{}l|c|c|c|c|cc|cc|cc|cc|cc|cc@{}}
\hline
 & & & & & \multicolumn{2}{c|}{\textbf{Set5}} &
  \multicolumn{2}{c|}{\textbf{Set14}} &
  \multicolumn{2}{c|}{\textbf{BSDS100}} &
  \multicolumn{2}{c|}{\textbf{Urban100}} &
  \multicolumn{2}{c|}{\textbf{Manga109}} & 
  \multicolumn{2}{c}{\textbf{ReSyn}}\\
\multirow{-2}{*}{Method} & \multirow{-2}{*}{scale} & \multirow{-2}{*}{\#param}& \multirow{-2}{*}{MACs}& \multirow{-2}{*}{dataset} & PSNR  & SSIM   & PSNR  & SSIM   & PSNR  & SSIM   & PSNR  & SSIM   & PSNR  & SSIM   & PSNR  & SSIM \\ \hline
SwinIR-light~\cite{liang2021swinir}& $\times$2  & 878K & {195.6G} & DIV2K %
& 37.76 & 0.9598 & 33.34 & 0.9161 & 32.05 & 0.8981 & 31.55 & 0.9225 & 38.08 & 0.9759 & 34.81 & 0.9276 
\\
SRFormer-light~\cite{zhou2023srformer}& $\times$2  & 853K & {236G} & DIV2K %
& 
37.68 & 0.9593 & 33.29 & 0.915 & 32.03 & 0.8977 & 31.56 & 0.9224 & 38.11 & 0.9763 & 34.93 & 0.9279
\\
MambaIR~\cite{guo2024mambair} & $\times$2  & 859K & {198.1G} & DIV2K %
 & 37.88 & 0.9601 & 33.50 & \best{0.9168} & \best{32.13} & \best{0.8993} & 31.97 & 0.9268 & 38.50 & 0.9767 & 34.96 & \best{0.9288}
\\
\best{RWKV-IR (Ours)} & $\times$2  & 863K & { 198.5G} & DIV2K %
& \best{37.98}
& \best{0.9604}
& \best{33.51}
& 0.9167
& 32.13
& 0.8991
& \best{32.15}
& \best{0.9283}
& \best{38.66}
& \best{0.9769}
& \best{34.96}
& 0.9287
\\
\hdashline
SwinIR-light~\cite{liang2021swinir}& $\times$2  & 878K & {195.6G} & ReSyn %
& 37.62 & 0.9594 & 33.29 & 0.9155 & 32.04 & 0.8980 & 31.54 & 0.9224 & 37.87 & 0.9760 & 34.92 & 0.9280 \\
SRFormer-light~\cite{zhou2023srformer}& $\times$2  & 853K & {236G} & ReSyn %
& 37.62 & 0.9594 & 33.29 & 0.9155 & 32.04 & 0.8980 & 31.54 & 0.9224 & 37.87 & 0.9760 & 34.92 & 0.9280 
\\
MambaIR~\cite{guo2024mambair} & $\times$2  & 859K & {198.1G} & ReSyn %
 & 37.68 & 0.9596 & 33.43 & 0.9162 & 32.10 & 0.8989 & 31.80 & 0.9251 & 38.52 & 0.9770 & 35.06 & \best{0.9290}
\\
\best{RWKV-IR (Ours)} & $\times$2  & 863K & { 198.5G} & ReSyn %
& \best{37.79} 
& \best{0.9601}
& \best{33.36}
& \best{0.9165}
& \best{32.09}  
& \best{0.8990}
& \best{31.98}
& \best{0.9263} 
& \best{38.65}
& \best{0.9775}
& \best{35.11}
& 0.9295
\\
\hline
SwinIR-light~\cite{liang2021swinir} & $\times$3  & 886K & {87.2G} & DIV2K %
& 34.11 & 0.9246 & 30.22 & 0.8399 & 28.97 & 0.8023 & 27.71 & 0.8426 & 33.00 & 0.9408 & 31.79 & 0.8687 
\\
SRFormer-light~\cite{zhou2023srformer} & $\times$3  & 861K & {105G} & DIV2K %
 & 34.24 & 0.9259 & 30.26 & 0.8408 & 29.01 & 0.8036 & 27.88 & 0.8474 & 33.21 & 0.9424 & 31.85 & 0.8699
\\ 
MambaIR~\cite{guo2024mambair} & $\times$3  & 867K & {88.7G} & DIV2K %
 & 34.32 & 0.9263 & \best{30.27} & 0.8406 & 29.05 & {0.8039} & 28.07 & 0.8506 & {33.39} & 0.9432 & \best{31.96} & \best{0.8711}
\\
\best{RWKV-IR (Ours)} & $\times$3  & 873K & { 91.7G} & DIV2K %
& \best{34.35} & \best{0.9265} & 30.20 & \best{0.8411} & \best{29.07} & \best{0.8044} & \best{28.13} & \best{0.8516} & \best{33.57} & \best{0.9439} & 31.95 & 0.8705
\\
\hdashline
SwinIR-light~\cite{liang2021swinir} & $\times$3  & 886K & {87.2G} & ReSyn %
& 34.07
& 0.9251
& 30.14
& 0.8385
& 28.96
& 0.8020
& 27.72
& 0.8427
& 32.87
& 0.9409
& 31.86
& 0.8692
\\
SRFormer-light~\cite{zhou2023srformer} & $\times$3  & 861K & {105G} & ReSyn %
 & 34.15 & 0.9255 & 30.19 & 0.8386 & 28.99 & 0.8021 & 27.82 & 0.8445 & 33.07 & 0.9418 & 31.91 & 0.8695
\\ 
MambaIR~\cite{guo2024mambair} & $\times$3  & 867K & {88.7G} & ReSyn %
& 34.20 & 0.9260 & \best{30.21} & 0.8391 & 29.04 & 0.8037 & 28.04 & 0.8490 & 33.29 & 0.9430 & 32.03 & 0.8713
\\
\best{RWKV-IR (Ours)} & $\times$3  & 873K & { 91.7G} & ReSyn %
& \best{34.29}
& \best{0.9264}
& 30.14
& \best{0.8392}
& \best{29.06}
& \best{0.8038}
& \best{28.15}
& \best{0.8505 }
& \best{33.46}
& \best{0.9439}
& \best{32.05}
& \best{0.8713}
\\
\hline
SwinIR-light~\cite{liang2021swinir} & $\times$4  & 897K & {49.6G} & DIV2K %
& 31.97 & 0.8913 & 28.47 & 0.7786 & 27.48 & 0.7336 & 25.76 & 0.7753 & 30.06 & 0.9031 & 30.11 & 0.8225 
\\
SRFormer-light~\cite{zhou2023srformer} & $\times$4  & 873K & {62.8G} & DIV2K %
 & 32.00 & 0.8915 & 28.46 & 0.7784 & 27.48 & 0.7337 & 25.84 & 0.7780 & 30.12 & 0.9039 & 30.11 & 0.8228
\\
MambaIR~\cite{guo2024mambair}  & $\times$4  &  879K & {50.6G} & DIV2K %
 & 32.12 & 0.8935 & \best{28.51} & 0.7796 & 27.51 & 0.7339 & 25.94 & 0.7804 & 30.27 & 0.9051 & 30.22 & 0.8241
\\
\best{RWKV-IR (Ours)} & $\times$4  & 887K & { 51.6G} & DIV2K %
& \best{32.14} & \best{0.8942} & 28.46 & \best{0.7809} & \best{27.57} & \best{0.7365} & \best{26.07} & \best{0.7848} & \best{30.48} & \best{0.9074} & \best{30.24} & \best{0.8246}
\\
\hdashline
SwinIR-light~\cite{liang2021swinir} & $\times$4  & 897K & {49.6G} & ReSyn %
& 31.93 & 0.8914 & 28.42 & 0.7778 & 27.45 & 0.7320 & 25.74 & 0.7742 & 30.04 & 0.9031 & 30.14 & 0.8222 
\\
SRFormer-light~\cite{zhou2023srformer} & $\times$4  & 873K & {62.8G} & ReSyn %
 & 31.97 & 0.8920 & 28.49 & 0.7782 & 27.47 & 0.7323 & 25.85 & 0.7771 & 30.16 & 0.9047 & 30.18 & 0.8230
\\
MambaIR~\cite{guo2024mambair}  & $\times$4  &  879K & {50.6G} & ReSyn %
 & 32.09 & 0.8938 & \best{28.53} & 0.7793 & 27.52 & 0.7337 & 26.02 & 0.7824 & 30.36 & 0.9070 & 30.31 & 0.8252
\\
\best{RWKV-IR (Ours)} & $\times$4  & 887K & { 51.6G} & ReSyn %
&   \best{32.16}   &  \best{0.8941}    &  28.45   &   \best{0.7806}  &   \best{27.56}  &  \best{0.7351}    &  \best{26.13}    &  \best{0.7850} & \best{30.47} & \best{0.9084}    &  \best{30.34}   &  \best{0.8251}  
\\
\hline
\end{tabular}%
}
\end{table*}

\begin{table*}[!t]
\centering
\caption{Quantitative comparison on \underline{\textbf{classic image super-resolution}} with state-of-the-art methods on 500K training iter{ation}s.}
\label{tab:classicSR_500K}
\vspace{1mm}
\renewcommand{\arraystretch}{1.0}
\setlength{\tabcolsep}{6pt}
\resizebox{1\linewidth}{!}{
\begin{tabular}{@{}l|c|cc|cc|cc|cc|cc@{}}
\hline
 & & \multicolumn{2}{c|}{\textbf{Set5}} &
  \multicolumn{2}{c|}{\textbf{Set14}} &
  \multicolumn{2}{c|}{\textbf{BSDS100}} &
  \multicolumn{2}{c|}{\textbf{Urban100}} &
  \multicolumn{2}{c}{\textbf{Manga109}} \\
\multirow{-2}{*}{Method} & \multirow{-2}{*}{scale} & PSNR  & SSIM   & PSNR  & SSIM   & PSNR  & SSIM   & PSNR  & SSIM   & PSNR  & SSIM   \\ \hline
EDSR~\cite{lim2017enhanced}   & $\times 2$ & 38.11 & 0.9602 & 33.92 & 0.9195 & 32.32 & 0.9013 & 32.93 & 0.9351 & 39.10 & 0.9773 \\
SAN~\cite{dai2019second}    & $\times 2$ & 38.31 & 0.9620 & 34.07 & 0.9213 & 32.42 & 0.9028 & 33.10 & 0.9370 & 39.32 & 0.9792 \\
HAN~\cite{niu2020single}    & $\times 2$ & 38.27 & 0.9614 & 34.16 & 0.9217 & 32.41 & 0.9027 & 33.35 & 0.9385 & 39.46 & 0.9785 \\
ELAN~\cite{zhang2022efficient}   & $\times 2$ & 38.36 & 0.9620 & 34.20 & 0.9228 & 32.45 & 0.9030 & 33.44 & 0.9391 & 39.62 & 0.9793 \\
SwinIR~\cite{liang2021swinir} & $\times 2$ & 38.42 & 0.9623 & 34.46 & 0.9250 & 32.53 & 0.9041 & 33.81 & 0.9427 & 39.92 & 0.9797 \\
SRFormer~\cite{zhou2023srformer} & $\times 2$ &  38.51 &0.9627&34.44&0.9253&32.57 &0.9046 &34.09& \best{0.9449} &40.07 &0.9802\\
MambaIR~\cite{guo2024mambair}   & $\times 2$ &\second{38.60}&\second{0.9628}&\best{34.69}&\best{0.9260}&\best{32.60}&\best{0.9048}&\second{34.17}&0.9443&\second{40.33}&\best{0.9806} \\ 
\best{RWKV-IR (Ours)}& $\times 2$ & \best{38.62} &\best{0.9629}&\second{34.63}&\second{0.9254}&\second{32.58} &\second{0.9045} &\best{34.20}& \second{0.9446} &\best{40.34} &\second{0.9804}\\
\hline
EDSR~\cite{lim2017enhanced} & $\times$3 & 
34.65 & 0.9280 & 30.52 & 0.8462 & 29.25 & 0.8093 & 28.80 & 0.8653 & 34.17 & 0.9476\\
SAN~\cite{dai2019second} & $\times$3 &
34.75 & 0.9300 & 30.59 & 0.8476 & 29.33 & 0.8112 & 28.93 & 0.8671 & 34.30 & 0.9494\\
HAN~\cite{niu2020single} & $\times$3 &
34.75 & 0.9299 & 30.67 & 0.8483 & 29.32 & 0.8110 & 29.10 & 0.8705 & 34.48 & 0.9500\\
ELAN~\cite{zhang2022efficient} & $\times$3 &
34.90 & 0.9313 & 30.80 & 0.8504 & 29.38 & 0.8124 & 29.32 & 0.8745 & 34.73 & 0.9517\\
SwinIR~\cite{liang2021swinir} & $\times$3 &
34.97 & 0.9318 & 30.93 & 0.8534 & 29.46 & 0.8145 & 29.75 & 0.8826 & 35.12 & 0.9537\\
SRformer~\cite{zhou2023srformer} & $\times$3 &
35.02& 0.9323 & 30.94 & \second{0.8540} & 29.48 & 0.8156 & \best{30.04} & \best{0.8865} & 35.26 & 0.9543\\
MambaIR~\cite{guo2024mambair} & $\times$3 &  \second{35.13} & \second{0.9326} & \best{31.06} & \best{0.8541} & \best{29.53} & \best{0.8162} & 29.98 & 0.8838 & \second{35.55} & \best{0.9549}\\ 
\best{RWKV-IR (Ours)}& $\times$3 &
\best{35.16}& \best{0.9328} & \second{31.02} & 0.8538 & \second{29.52} & \second{0.8159} & \second{30.02} & \second{0.8839} & \best{35.57} & \second{0.9548}\\
\hline
EDSR~\cite{lim2017enhanced}   & $\times 4$ & 32.46 & 0.8968 & 28.80 & 0.7876 & 27.71 & 0.7420 & 26.64 & 0.8033 & 31.02 & 0.9148 \\
SAN~\cite{dai2019second}      & $\times 4$ & 32.64 & 0.9003 & 28.92 & 0.7888 & 27.78 & 0.7436 & 26.79 & 0.8068 & 31.18 & 0.9169 \\
HAN~\cite{niu2020single}      & $\times 4$ & 32.64 & 0.9002 & 28.90 & 0.7890 & 27.80 & 0.7442 & 26.85 & 0.8094 & 31.42 & 0.9177 \\
ELAN~\cite{zhang2022efficient}   & $\times 4$ & 32.75 & 0.9022 & 28.96 & 0.7914 & 27.83 & 0.7459 & 27.13 & 0.8167 & 31.68 & 0.9226 \\
SwinIR~\cite{liang2021swinir}    & $\times 4$ & 32.92 & 0.9044 & 29.09 & 0.7950 & 27.92 & 0.7489 & 27.45 & 0.8254 & 32.03 & 0.9260 \\
SRFormer~\cite{zhou2023srformer} & $\times 4$ & 32.93 & 0.9041 & 29.08 &0.7953 &27.94 & 0.7502 & 27.68 & \best{0.8311} & 32.21 & 0.9271 \\
MambaIR~\cite{guo2024mambair}   & $\times 4$ &\second{33.13}&\second{0.9054}&\best{29.25}&\best{0.7971}&\best{28.01}&\second{0.7510}&\second{27.80}&{0.8303}&\second{32.48}&\second{0.9281} \\  
\best{RWKV-IR (Ours)}& $\times 4$  &\best{33.14}& \best{0.9056} & \second{29.20} &\second{0.7968} &\second{27.99} & \best{0.7511} & \best{27.83} & \second{0.8305} & \best{32.51} & \best{0.9285} \\
\hline
\end{tabular}%
}
\vspace{0mm}
\end{table*}

\subsection{Lightweight Image Super-Resolution} We also provide comparison of our RWKV-IR-light with state-of-the-art lightweight image SR methods: SwinIR~\cite{liang2021swinir}, SRFormer~\cite{zhou2023srformer} and MambaIR~\cite{guo2024mambair}. Including PSNR and SSIM, we also compare the total number of parameters and MACs (multiply-accumulate operations) to show the model size and the computational complexity of different models. We compare the metrics gained from different training datasets used. The results on 50K training {iterations} are shown in Tab.~\ref{tab:lightSR} and on 500K are shown in Tab.~\ref{tab:lightSR_500K}. 
On {the small number of} training {iterations}, RWKV-IR outperforms MambaIR-light by 0.10dB on Urban100 with an x4 scale{,} with a similar parameter number and MACs when trained on the DIV2K dataset. Using ReSyn get{s} a similar result with DIV2K, since {the} lightweight models' training iter{ation} is small, our ReSyn could also show a favorable performance. 
On the large number of training iter{ation}s, our proposed linear-complexity attention-based image restoration model also shows competitive performance. This indicates that our model not only converges quickly{,} but also has excellent image restoration capabilities.

\begin{table*}[t]
\centering
\caption{Quantitative comparison on \underline{\textbf{lightweight image super-resolution}} with state-of-the-art methods on 500K training iter{ation}s. }
\label{tab:lightSR_500K}
\vspace{1mm}
\renewcommand{\arraystretch}{1.0}
\setlength{\tabcolsep}{6pt}
\resizebox{1\linewidth}{!}{
\begin{tabular}{@{}l|c|c|c|cc|cc|cc|cc|cc@{}}
\hline
 & & & & \multicolumn{2}{c|}{\textbf{Set5}} &
  \multicolumn{2}{c|}{\textbf{Set14}} &
  \multicolumn{2}{c|}{\textbf{BSDS100}} &
  \multicolumn{2}{c|}{\textbf{Urban100}} &
  \multicolumn{2}{c}{\textbf{Manga109}} \\
\multirow{-2}{*}{Method} & \multirow{-2}{*}{scale}& \multirow{-2}{*}{\#param}& \multirow{-2}{*}{MACs} & PSNR  & SSIM   & PSNR  & SSIM   & PSNR  & SSIM   & PSNR  & SSIM   & PSNR  & SSIM   \\ \hline
CARN~\cite{ahn2018fast} & $\times$2  & 1,592K & 222.8G
& 37.76
& 0.9590
& 33.52
& 0.9166
& 32.09
& 0.8978
& 31.92
& 0.9256
& 38.36
& 0.9765
\\
IMDN~\cite{hui2019lightweight}& $\times$2  & 694K & 158.8G
& 38.00
& 0.9605
& 33.63
& 0.9177
& 32.19
& 0.8996
& 32.17
& 0.9283
& {38.88}
& {0.9774}
\\
LAPAR-A~\cite{li2020lapar} & $\times$2  & 548K & 171.0G
& 38.01
& 0.9605
& 33.62
& 0.9183
& 32.19
& 0.8999
& 32.10
& 0.9283
& 38.67
& 0.9772
\\
SwinIR-light~\cite{liang2021swinir}& $\times$2  & 878K & {195.6G} %
& {38.14}
& {0.9611}
& {33.86}
& {0.9206}
& {32.31}
& {0.9012}
& {32.76}
& {0.9340}
& {39.12}
& {0.9783}
\\
SRFormer-light~\cite{zhou2023srformer}& $\times$2  & 853K & {236G} %
& \best{38.23}
& {0.9613}
& {33.94}
& {0.9209}
& {32.36}
& \best{0.9019}
& {32.91}
& {0.9353}
& {39.28}
& \best{0.9785}
\\
MambaIR~\cite{guo2024mambair} & $\times$2  & 859K & {198.1G} %
& {38.16}
& {0.9610}
& \best{34.00}
& \best{0.9212}
& {32.34}
& {0.9017}
& {32.92}
& {0.9356}
& {39.31}
& {0.9779}
\\
\best{RWKV-IR (Ours)} & $\times$2  & 859K & {198.1G} %
& {38.22}
& \best{0.9614}
& {33.98}
& {0.9210}
& \best{32.37}
& {0.9018}
& \best{32.95}
& \best{0.9359}
& \best{39.34}
& {0.9781}
\\
\hline

CARN~\cite{ahn2018fast} & $\times$3  & 1,592K  & 118.8G
& 34.29
& 0.9255
& 30.29
& 0.8407
& 29.06
& 0.8034
& 28.06
& 0.8493
& 33.50 
& 0.9440
\\ 
IMDN~\cite{hui2019lightweight} & $\times$3  & 703K  & 71.5G 
& 34.36
& 0.9270
& 30.32
& 0.8417
& 29.09
& 0.8046
& 28.17
& 0.8519
& {33.61}
& {0.9445}
\\ 
LAPAR-A~\cite{li2020lapar} & $\times$3  & 544K & 114.0G
& 34.36
& 0.9267
& 30.34
& 0.8421
& 29.11
& 0.8054
& 28.15
& 0.8523
& 33.51
& 0.9441
\\
SwinIR-light~\cite{liang2021swinir} & $\times$3  & 886K & {87.2G} %
& {34.62}
& {0.9289}
& {30.54}
& {0.8463}
& {29.20}
& {0.8082}
& {28.66}
& {0.8624}
& {33.98}
& {0.9478}
\\
SRFormer-light~\cite{zhou2023srformer} & $\times$3  & 861K & {105G} %
& {34.67}
& {0.9296}
& {30.57}
& {0.8469}
& {29.26}
& {0.8099}
& {28.81}
& {0.8655}
& {34.19}
& {0.9489}
\\ 
MambaIR~\cite{guo2024mambair} & $\times$3  & 867K & {88.7G} %
& {34.72}
& {0.9296}
& \best{30.63}
& \best{0.8475}
& {29.29}
& \best{0.8099}
& {29.00}
& {0.8689}
& \best{34.39}
& \best{0.9495}
\\
\best{RWKV-IR (Ours)} & $\times$3  & 867K & {88.7G} %
& \best{34.76}
& \best{0.9301}
& {30.59}
& {0.8471}
& \best{29.32}
& {0.8096}
& \best{29.04}
& \best{0.8693}
& {34.37}
& {0.9491}
\\
\hline

CARN~\cite{ahn2018fast} & $\times$4  & 1,592K & 90.9G
& 32.13
& 0.8937
& 28.60
& 0.7806
& 27.58
& 0.7349
& 26.07 
& 0.7837
& {30.47}
& {0.9084}
\\
IMDN~\cite{hui2019lightweight}& $\times$4  & 715K & 40.9G
& 32.21
& 0.8948
& 28.58
& 0.7811
& 27.56
& 0.7353 
& 26.04
& 0.7838
& 30.45
& 0.9075
\\
LAPAR-A~\cite{li2020lapar} & $\times$4  & 659K & 94.0G
& 32.15
& 0.8944
& 28.61
& 0.7818
& 27.61
& 0.7366
& 26.14
& 0.7871
& 30.42
& 0.9074
\\
SwinIR-light~\cite{liang2021swinir} & $\times$4  & 897K & {49.6G} %
& {32.44}
& {0.8976}
& {28.77}
& {0.7858}
& {27.69}
& {0.7406}
& {26.47}
& {0.7980}
& {30.92}
& {0.9151}
\\
SRFormer-light~\cite{zhou2023srformer} & $\times$4  & 873K & {62.8G} %
& {32.51}
& {0.8988}
& {28.82}
& {0.7872}
& {27.73}
& {0.7422}
& {26.67}
& {0.8032}
& {31.17}
& {0.9165}
\\
MambaIR~\cite{guo2024mambair}  & $\times$4  &  879K & {50.6G} %
& {32.51}
& {0.8993}
& \best{28.85}
& \best{0.7876}
& {27.75}
& {0.7423}
& {26.75}
& {0.8051}
& {31.26}
& {0.9175}
\\
\best{RWKV-IR (Ours)}  & $\times$4  &  879K & {50.6G} %
& \best{32.53}
& \best{0.8995}
& {28.82}
& {0.7875}
& \best{27.78}
& \best{0.7426}
& \best{26.79}
& \best{0.8052}
& \best{31.28}
& \best{0.9179}
\\
\hline
\end{tabular}
}
\end{table*}

\subsection{Gaussian Color Image Denoising} 

As shown in Tab.~\ref{tab:guassian-denoise}, we conduct the quantitative comparison between our RWKV-IR and the SOTA methods IRCNN~\cite{IRCNN},  FFDNet~\cite{FFDNet}, DnCNN~\cite{DnCNN}, SwinIR~\cite{liang2021swinir}, Restormer~\cite{zamir2022restormer} and MambaIR~\cite{guo2024mambair} on long training ite{ration}s. All the models are trained on the DFWB-RGB dataset. Our method {achieves} competitive metrics on all four datasets.

\begin{table*}[!t]
\centering
\caption{Quantitative comparison on \underline{\textbf{gaussian color image denoising}} with state-of-the-art methods.}
\label{tab:guassian-denoise}
\vspace{1mm}
\renewcommand{\arraystretch}{1.0}
\setlength{\tabcolsep}{10pt}
\resizebox{1\linewidth}{!}{
\begin{tabular}{l|ccc|ccc|ccc|ccc}
\hline
\multirow{2}{*}{Method} &  \multicolumn{3}{c|}{\textbf{BSD68}} & \multicolumn{3}{c|}{\textbf{Kodak24}} & \multicolumn{3}{c|}{\textbf{McMaster}} & \multicolumn{3}{c}{\textbf{Urban100}}\\
&$\sigma$=15 & $\sigma$=25 & $\sigma$=50 &$\sigma$=15 & $\sigma$=25 & $\sigma$=50 &$\sigma$=15 & $\sigma$=25 & $\sigma$=50 &$\sigma$=15 & $\sigma$=25 & $\sigma$=50\\
\hline
IRCNN~\cite{IRCNN} &
33.86 & 31.16 & 27.86 & 34.69 & 32.18 & 28.93 & 34.58 & 32.18 & 28.91 & 33.78 & 31.20 & 27.70\\
FFDNet~\cite{FFDNet} &
33.87 & 31.21 & 27.96 & 34.63 & 32.13 & 28.98 & 34.66 & 32.35 & 29.18 & 33.83 & 31.40 & 28.05\\
DnCNN~\cite{DnCNN} &
33.90 & 31.24 & 27.95 & 34.60 & 32.14 & 28.95 & 33.45 & 31.52 & 28.62 & 32.98 & 30.81 & 27.59\\
DRUNet~\cite{DRUNet}&
34.30 & 31.69 & 28.51 & 35.31 & 32.89 & 29.86 & 35.40 & 33.14 & 30.08 & 34.81 & 32.60 & 29.61\\
SwinIR~\cite{liang2021swinir} &
34.42 & 31.78 & 28.56 & 35.34 & 32.89 & 29.79 & 35.61 & 33.20 & 30.22 & 35.13 & 32.90 & 29.82\\
Restormer~\cite{zamir2022restormer} &
34.40 & 31.79 & 28.60 & \best{35.47} & \best{33.04} & \best{30.01} & 35.61 & 33.34 & {30.30} & 35.13 & 32.96 & 30.02\\ 
MambaIR~\cite{guo2024mambair} &
\best{34.44} & \best{31.82} & \best{28.64} & {35.35} & {32.92} & {29.87} & \best{35.63} &\best{33.36} & \second{30.32} & \second{35.17} & \second{32.99} & \second{30.06} \\
\best{RWKV-IR (Ours)} &
\second{34.43} & \second{31.79} & \second{28.62} & \second{35.37} & \second{32.98} & \second{29.92} & \second{35.62} &\second{33.35} & \best{30.33} & \best{35.19} & \best{33.02} &\best {30.10} \\
\hline
\end{tabular}
}
\end{table*}

\subsection{Grayscale Image Denoising} 

As shown in Tab.~\ref{tab:gray-denoise}, we conduct the quantitative comparison between our RWKV-IR and the SOTA methods IRCNN~\cite{IRCNN}, FFDNet~\cite{FFDNet}, DnCNN~\cite{DnCNN}, SwinIR~\cite{liang2021swinir} 
on grayscale image denoising task. All models are trained on long iter{ation}s of 500K and on DFWB-gray dataset. Our method {achieves} competitive metrics on all three datasets.

\begin{table*}[!t]
\centering
\caption{Quantitative comparison on \underline{\textbf{grayscale image denoising}} with state-of-the-art methods.}
\label{tab:gray-denoise}
\vspace{1mm}
\renewcommand{\arraystretch}{1.0}
\setlength{\tabcolsep}{12pt}
\resizebox{1\linewidth}{!}{
\begin{tabular}{l|ccc|ccc|ccc}
\hline
\multirow{2}{*}{Method} &  \multicolumn{3}{c|}{\textbf{Set12}} & \multicolumn{3}{c|}{\textbf{BSD68}} & \multicolumn{3}{c}{\textbf{Urban100}} \\
&$\sigma$=15 & $\sigma$=25 & $\sigma$=50 &$\sigma$=15 & $\sigma$=25 & $\sigma$=50 &$\sigma$=15 & $\sigma$=25 & $\sigma$=50 \\
\hline
IRCNN~\cite{IRCNN} &
32.76 & 30.37 & 27.12 & 31.63 & 29.15 & 26.19 & 32.46 & 29.80 & 26.22 \\
FFDNet~\cite{FFDNet} &
32.75 & 30.43 & 27.32 & 31.63 & 29.19 & 26.29 & 32.40 & 29.90 & 26.50 \\
DnCNN~\cite{DnCNN} &
33.86 & 30.44 & 27.18 & 31.73 & 29.23 & 26.23 & 32.64 & 29.95 & 26.26 \\
DRUNet~\cite{DRUNet}&
33.25 & 30.94 & 27.90 & 31.91 & 29.48 & 26.59 & 33.44 & 31.11 & 27.96 \\
SwinIR~\cite{liang2021swinir} &
33.36 & 31.01 & 27.91 & 31.97 & 29.50 & 26.58 & 33.70 & 31.30 & 27.98 \\
MambaIR~\cite{guo2024mambair} &
\second{34.44} & \second{31.82} & \second{28.64} & \best{35.35} & \best{32.92} & \best{29.87} & \second{35.63} & \best{33.36} & \second{30.32} \\
\best{RWKV-IR (Ours)} &
\best{34.46} & \best{31.85} & \best{28.66} & \second{35.33} & \second{32.90} & \second{29.84} & \best{35.64} &\second{33.35} & \best{30.34} \\
\hline
\end{tabular}
}
\end{table*}

\begin{table*}[!t]
\centering
\caption{Quantitative comparison on \underline{\textbf{JPEG compression artifact reduction}} with state-of-the-art methods. {We show scores of average PSNR/SSIM/PSNR-B.}}
\label{tab:JPEG}
\vspace{1mm}
\renewcommand{\arraystretch}{1.0}
\setlength{\tabcolsep}{2pt}
\resizebox{1\linewidth}{!}{
\begin{tabular}{l|cccc|cccc}
\hline
\multirow{2}{*}{ Method} &  \multicolumn{4}{c|}{\textbf{Classic5}} & \multicolumn{4}{c}{\textbf{LIVE1}} \\
&$q$=10 & $q$=20 & $q$=30 &$q$=40 & $q$=10 & $q$=20&$q$=30 & $q$=40 \\
\hline 
ARCNN~\cite{zhang2018rcan} &
29.03/0.7929/28.76 & 31.15/0.8517/30.59 & 32.51/0.8806/31.98 & 33.32/0.8953/32.79 &  28.96/0.8076/28.77 & 31.29/0.8733/30.79 & 32.67/0.9043/32.22 & 33.63/0.9198/33.14 \\
DnCNN-3~\cite{DnCNN} &
29.40/0.8026/29.13 &  31.63/0.8610/31.19 &  31.63/0.8610/31.19 & 33.77/0.9003/33.20 & 29.19/0.8123/28.90 & 31.59/0.8802/31.07 & 32.98/0.9090/32.34 &  33.96/0.9247/33.28 \\
DRUNet~\cite{DRUNet}&
30.16/0.8234/29.81 & 32.39/0.8734/31.80 & 33.59/0.8949/32.82 & 34.41/0.9075/33.51 & 29.79/0.8278/29.48 & 32.17/0.8899/31.69 & 33.59/0.9166/32.99 & 34.58/0.9312/33.93  \\
SwinIR~\cite{liang2021swinir} &
30.27/0.8249/29.95 & 32.52/0.8748/31.99 & 33.73/0.8961/33.03 & 34.52/0.9082/33.66 & 29.86/0.8287/29.50 & 32.25/0.8909/31.70 & 33.69/0.9174/33.01 & 34.67/0.9317/33.88  \\ 
\best{RWKV-IR (Ours)} &
\best{30.35/0.8261/30.04} & \best{32.63/0.8760/32.05} & \best{33.81/0.8972/33.12} & \best{34.61/0.9091/33.71} & \best{29.94/0.8296/29.62} & \best{32.34/0.8915/31.81} & \best{33.78/0.9185/33.12} &\best{34.78/0.9323/33.95}  \\
\hline
\end{tabular}
}
\end{table*}

\subsection{JPEG Compression Artifact Reduction}

Tab.~\ref{tab:JPEG} shows the comparison of RWKV-IR with state-of-the-art JPEG compression artifact reduction methods: ARCNN~\cite{zhang2018rcan}, DnCNN-3 ~\cite{DnCNN}, DRUNet~\cite{DRUNet} and SwinIR~\cite{liang2021swinir}. Following ~\cite{DRUNet, liang2021swinir}, we test different methods on two benchmark datasets (Classic5 and LIVE1) for JPEG quality factors 10, 20, 30 and 40. {It can be seen that} the proposed RWKV-IR has average PSNR gains of at least 0.07dB and 0.08dB on two testing datasets for different quality factors.


\end{document}